\documentclass[11pt]{article}

\usepackage[preprint]{acl}

\usepackage{times}
\usepackage{latexsym}

\usepackage[T1]{fontenc}
\usepackage[utf8]{inputenc}

\usepackage{microtype}

\usepackage{inconsolata}

\usepackage{graphicx}
\usepackage{amsmath}
\usepackage{amssymb}
\usepackage{algorithm}
\usepackage{algorithmic}
\usepackage{booktabs}
\usepackage{multirow}
\usepackage{array}
\usepackage{pifont}  
\usepackage{url}
\usepackage{hyperref}
\usepackage{placeins}
\title{DocIntent: Answerability-Guided Agentic Restoration for Real-World Document Visual Question Answering}

\author{\textbf{Zihan Huang, Shihang Wu, Junle Liu, Peirong Zhang} \\
  \textbf{Yongxin Shi, Xuhan Zheng, Lianwen Jin}\thanks{Corresponding author.} \\
  South China University of Technology}

\begin{document}
\maketitle

\begin{abstract}
Real-world degradations such as blur, shadow, distortion, and moir\'e patterns severely impair the document question-answering capabilities of Multimodal Large Language Models (MLLMs). Applying restoration tools before Visual Question Answering (VQA) is an intuitive solution. However, existing restoration approaches remain limited, as manually designing and executing restoration strategies is labor-intensive and requires domain expertise. Agentic restoration offers new possibilities for automation, yet existing frameworks primarily target natural images and pursue perceptual quality, overlooking that restoration should serve downstream tasks rather than optimize generic image quality metrics. To this end, we explore the value of agentic restoration for real-world degraded document VQA and propose DocIntent, a training-free Answerability-Guided Agentic Restoration framework. DocIntent first assesses question answerability, then identifies task-relevant degradations and selectively invokes restoration tools. A Comparison-Based Rollback mechanism validates each restoration step and reverts it when question-relevant evidence becomes less decipherable. The entire process requires no additional pretrained degradation classifier or image quality assessment model. Extensive experiments on the WildDoc benchmark show that DocIntent consistently improves the average score and consistency of different open- and closed-source MLLMs. The code and experimental data will be publicly available.
\end{abstract}

\section{Introduction}
\label{sec:introduction}
In the digital era, enterprises and organizations handle large volumes of documents, such as invoices, contracts, and forms~\cite{jaume2019funsd,park2019cord}. These documents often contain critical information for business workflows, making efficient and accurate information extraction essential. However, traditional manual processing is labor-intensive and time-consuming. Visual Question Answering (VQA) on document-related data addresses this challenge by enabling intelligent systems to understand document content and answer user-specified questions~\cite{mathew2021docvqa,mathew2022infographicvqa}, supporting more flexible document processing.

\begin{figure}[t]
  \centering
  \includegraphics[width=\columnwidth]{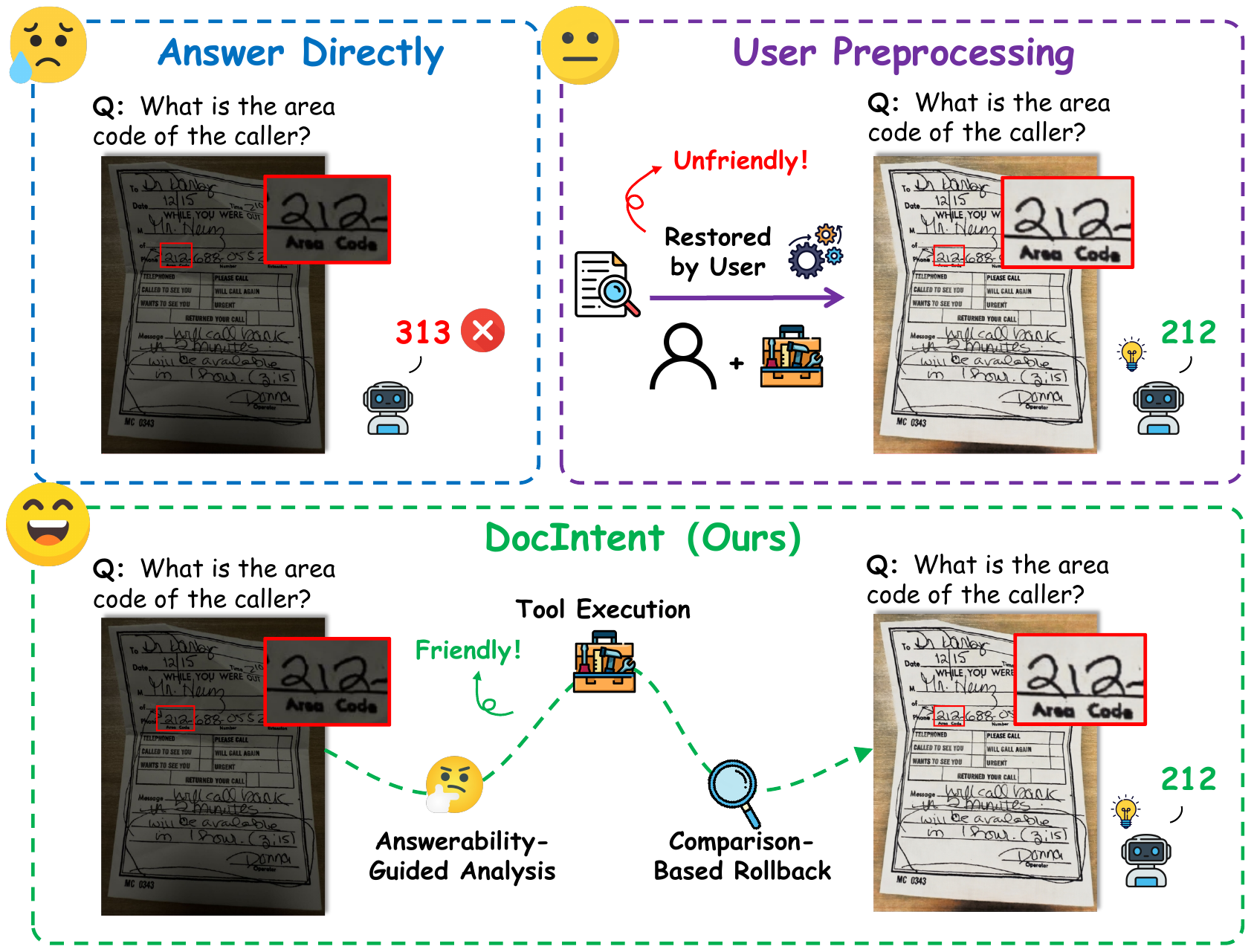}
  \caption{Comparison between existing real-world document VQA methods and the proposed DocIntent (bottom). The ground truth answer to the question is 212.}
  \label{fig:motivate}
\end{figure}

Recently, Multimodal Large Language Models (MLLMs)~\cite{qwen2.5-VL, team2023gemini,hurst2024gpt} have demonstrated remarkable potential in document understanding~\cite{mathew2021docvqa,mathew2022infographicvqa,masry2022chartqa}. However, real-world document images often exhibit degradations such as blur, shadow, and distortion~\cite{wang2025wilddoc,zhang2024docres,feng2021doctr,ocrgenbench2025zhang}, which severely hinder information extraction by MLLMs. Research on benchmarks such as WildDoc~\cite{wang2025wilddoc} reveals that even state-of-the-art MLLMs experience substantial performance degradation when confronted with degraded documents, as illustrated in Figure~\ref{fig:motivate}(top-left). A straightforward strategy is to preprocess degraded documents with restoration tools before feeding them into MLLMs (Figure~\ref{fig:motivate}, top-right). However, this pipeline depends on manually diagnosing degradation types and selecting appropriate tools, making it impractical for non-experts. Recently, agentic image restoration methods~\cite{chen2024restoreagent,zhou2025q,zhu2024intelligent, zuo20254kagent, zhang2026tir} have achieved success in natural image restoration. However, these methods primarily focus on natural image processing and pursue the highest perceptual image quality rather than aligning with downstream task requirements. In document question answering, different questions exhibit varying sensitivity to degradations, and not all degradations hinder question answering. Existing document restoration tools~\cite{zhang2024docres,yang2025unidemoire,zhao2025uni} optimize perceptual quality metrics such as SSIM~\cite{wang2004image} and PSNR. However, these overall quality metrics cannot ensure local text clarity. The restoration process may introduce artifacts, text blurring, or distortion that impair question-answering performance. Processing task-irrelevant degradations unnecessarily accumulates such risks. Therefore, a critical question arises: \textbf{How to tailor document restoration to specific question-answering tasks?}

To address this question, we propose \textbf{DocIntent}, a training-free and user-friendly answerability-guided document image restoration framework. As illustrated in Figure~\ref{fig:motivate} (bottom), DocIntent integrates downstream question awareness into restoration planning through a two-phase cyclic mechanism. In Phase A: Answerability-Guided Restoration (AGRes), the MLLM first assesses whether the current image provides sufficient information to answer the question. If restoration is needed, it identifies the degradation that most affects answerability and invokes the corresponding restoration tool to improve the document image. In Phase B: Comparison-Based Review, the Comparison-Based Rollback (CBRb) mechanism compares the restored image with the previous image to determine whether the operation improves the conditions for answering. Beneficial restorations are accepted for subsequent reasoning, while harmful ones are rolled back, and the corresponding tool is disabled. Both phases are performed by the MLLM without requiring an additional pretrained degradation classifier or image quality assessment model. Through this cycle, DocIntent selectively applies restoration to improve answerability while avoiding unnecessary or harmful preprocessing.

\begin{figure*}[t]
  \centering
  \includegraphics[width=\textwidth]{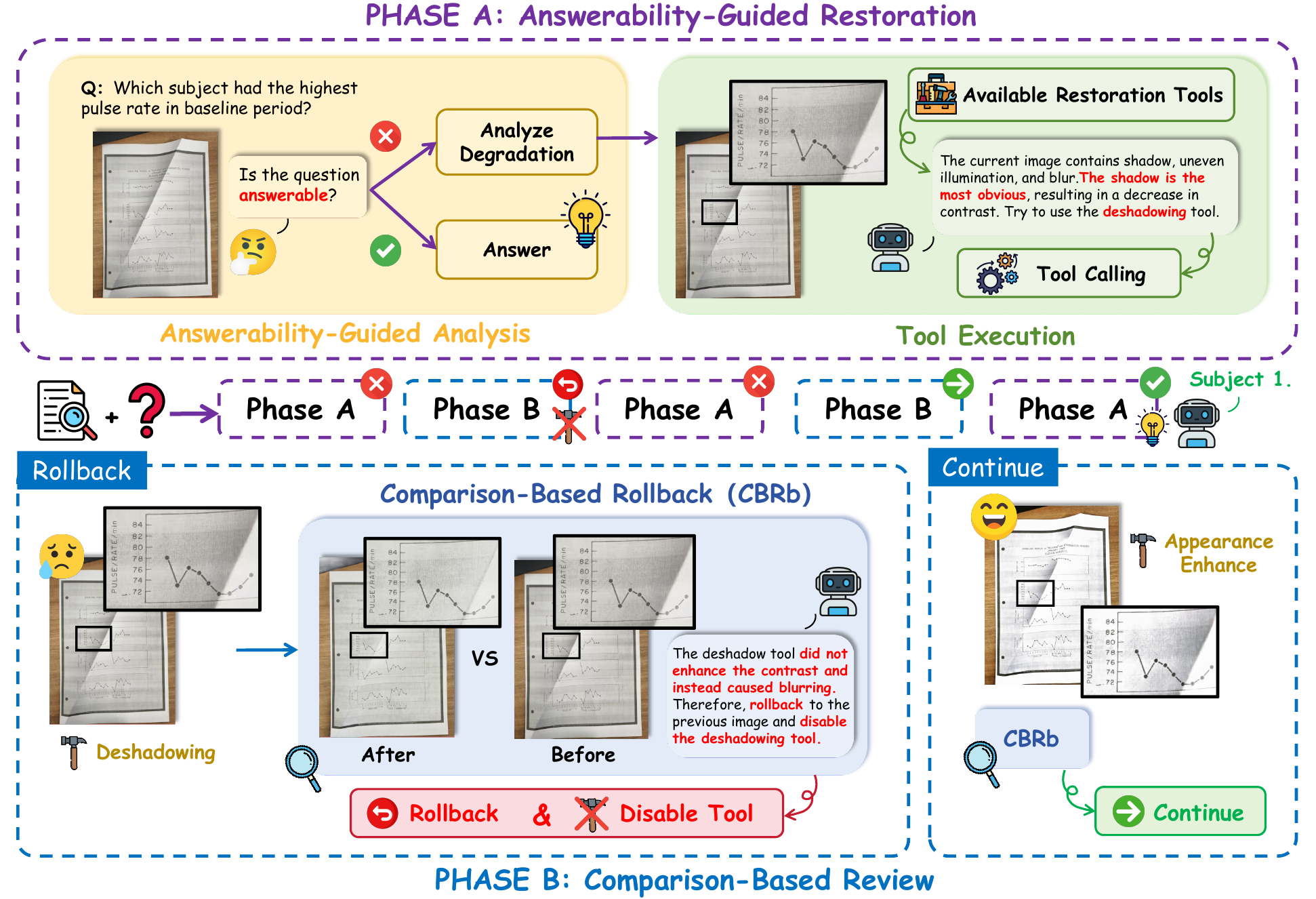}
  \caption{Overview of the \textbf{DocIntent} framework with two-phase cyclic mechanism: Phase A (Answerability-Guided Restoration) and Phase B (Comparison-Based Review). The figure illustrates an example where deshadowing triggers rollback (bottom-left), while appearance enhancement succeeds and continues (bottom-right). \label{fig:system}}
\end{figure*}

In summary, our contributions are as follows:
\begin{itemize}
  \item To the best of our knowledge, we are the first to explore the value of agentic restoration for document restoration and real-world degraded document VQA. We propose an Answerability-Guided Agentic Restoration framework that incorporates downstream question awareness into restoration planning.
  \item We design a two-phase cyclic mechanism with Answerability-Guided Restoration and Comparison-Based Rollback, enabling MLLMs to selectively invoke document restoration tools and roll back harmful results without additional training, a pretrained degradation classifier, or an image quality assessment model. A standardized tool interface also supports the integration of stronger restoration models in the future.
  \item Extensive experiments on the WildDoc~\cite{wang2025wilddoc} benchmark demonstrate that DocIntent consistently improves models' question-answering performance on degraded documents across different capability levels and achieves results close to human expert preprocessing.
\end{itemize}

\section{Related work}
\label{sec:related}

\subsection{Document Visual Question Answering}
Visual Question Answering (VQA) on document-related data~\cite{mathew2021docvqa, nandi2026document} requires models to answer questions about document images by understanding the text, images, and layout. Early methods~\cite{xu2021layoutlmv2, huang2022layoutlmv3, appalaraju2024docformerv2, zhu2025simple} adopted OCR-based pipelines, extracting textual and layout information \cite{lggpt2025zhang} through OCR systems before answering questions. Subsequent end-to-end approaches~\cite{kim2021donut, zhang2025docrouter, souibgui2025docvxqa} eliminated the OCR module, directly learning document representations through visual encoders. With the rise of MLLMs~\cite{qwen3.5, wang2025internvl3_5, team2023gemini,hurst2024gpt,anthropic2025claude4.6sonnet,seedseed2}, general-purpose MLLMs have demonstrated strong performance in document understanding tasks. Specialized document-understanding MLLMs~\cite{hu2025mplug, li2024monkey,huang2024mini, zhao2024harmonizing, yu2025minicpmv45cookingefficient} have also achieved significant results on standard benchmarks~\cite{mathew2021docvqa, masry2022chartqa, kim2024tablevqa, mathew2022infographicvqa}. Uni-DocRobust~\cite{zhouuni} improves robustness through feature restoration, whereas DocIntent selectively invokes image restoration tools based on question answerability.

However, existing MLLMs exhibit significant performance degradation on real-world document question answering~\cite{wang2025wilddoc}. Real-world degradations severely impair MLLMs' ability to perceive and understand document content. To address this challenge, we propose DocIntent, a training-free answerability-guided document restoration framework. DocIntent leverages MLLMs' inherent degradation recognition capability and employs a selective restoration strategy to enhance their document-question-answering performance under real-world degradation scenarios.

\begin{figure*}[t]
  \centering
  \includegraphics[width=\textwidth]{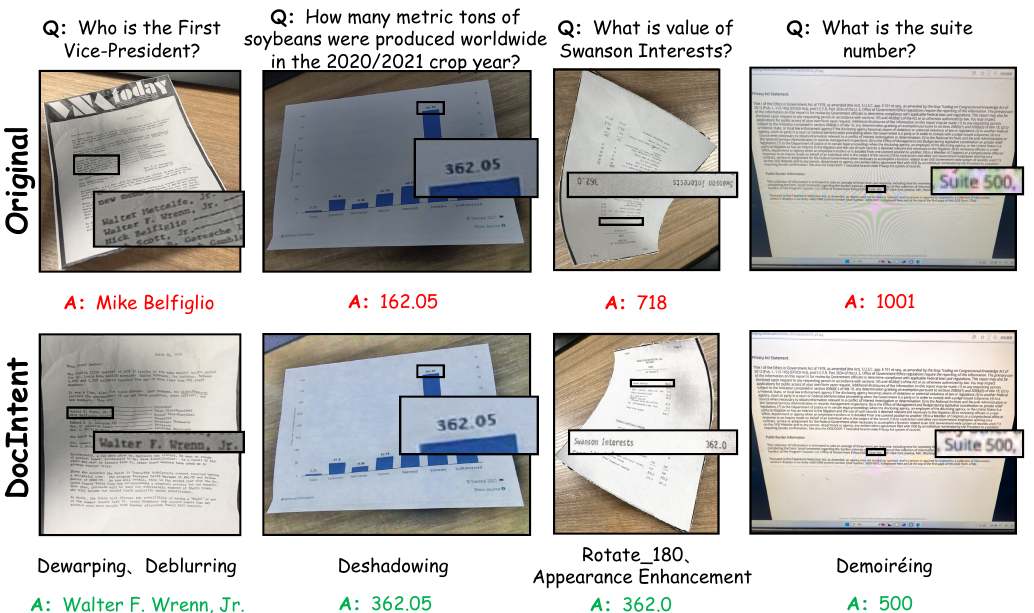}
  \caption{Cases where models fail on degraded images but succeed with DocIntent. Left two columns: Qwen3.5-9B~\cite{qwen3.5}. Right two columns: Gemini-3.1-Pro~\cite{gemini_3_1_pro_web}.}
  \label{fig:main_case}
\end{figure*}

\subsection{Agentic Image Restoration}

In the field of image restoration, RestoreAgent~\cite{chen2024restoreagent} pioneered the agentic image restoration paradigm, using task-specific trained models~\cite{zamir2022restormer} to restore images of natural scenes with multiple degradations. Existing agentic image restoration frameworks~\cite{zhu2024intelligent, chen2024restoreagent, li2025hybrid, zhou2025q, zuo20254kagent, zhang2026tir} primarily focus on natural images and pursue optimal quality. AgenticIR~\cite{zhu2024intelligent} uses a fine-tuned vision model for degradation recognition and quality assessment, while 4KAgent~\cite{zuo20254kagent} combines multiple image quality assessment models to plan and validate restoration results. In contrast, document image restoration is intended to serve specific downstream tasks rather than pursue the highest image quality. Different models and questions also exhibit varying sensitivities to degradations. Therefore, our approach does not rely on a specialized degradation classifier or image quality assessment model and requires no additional training. Instead, it enables MLLMs to adaptively assess whether the current image quality hinders answering the specific question at hand. Meanwhile, the emergence of multi-task document restoration models~\cite{zhang2024docres, yang2025unidemoire,zhao2025uni,peng2024upocr,docaligner2026zhang} provides strong support for the use of restoration tools.

Based on these observations, we propose DocIntent, an answerability-guided document image restoration framework. DocIntent adopts a selective restoration strategy, processing degradations that affect downstream tasks, and decides whether to accept restoration results through a comparison-based rollback mechanism, ensuring task performance while avoiding unnecessary processing overhead and risk of text quality degradation.

\section{Method}
\label{sec:method}

We first introduce the DocIntent framework, including three core modules: Answerability-Guided Analysis, Tool Execution, and Comparison-Based Rollback (Sec.~\ref{sec:framework}). We then detail the complete two-phase cyclic workflow, demonstrating how these modules collaborate to achieve answerability-guided selective restoration while preventing quality degradation (Sec.~\ref{sec:workflow}).

\subsection{DocIntent Framework}
\label{sec:framework}
Question answering on degraded document images involves several practical challenges, including determining whether restoration is needed, selecting appropriate restoration tools, and avoiding performance degradation caused by unsuitable restoration. To address these issues, DocIntent is designed with three core modules: Answerability-Guided Analysis, Tool Execution, and Comparison-Based Rollback, as shown in Figure~\ref{fig:system}.

\begin{figure}[t]
  \centering
  \includegraphics[width=\columnwidth]{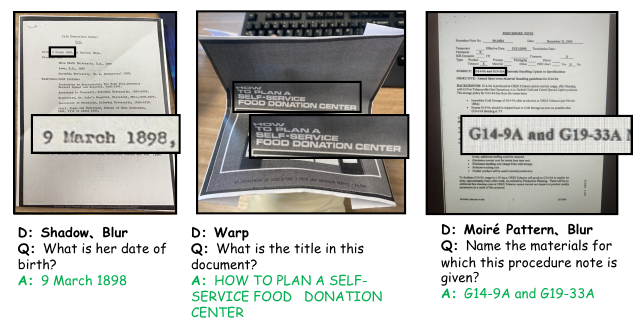}
  \caption{Representative cases where questions can be correctly answered despite visible degradations. D denotes degradation type.}
  \label{fig:answer_compare}
\end{figure}

\paragraph{Answerability-guided analysis.}
The MLLM first analyzes the image in conjunction with the question to determine whether the question can be answered directly from the current image. If so, it generates the answer directly without triggering restoration. Otherwise, the MLLM identifies which degradations prevent it from answering and selects the most severe one whose corresponding tool is available for restoration. Unlike traditional methods~\cite{zhu2024intelligent, chen2024restoreagent, li2025hybrid, zhou2025q} that pursue the highest image quality, our approach triggers restoration only when degradation hinders answering. As illustrated in Figure~\ref{fig:answer_compare}, when degradations do not affect task-critical regions, questions can be correctly answered without any restoration, making blind restoration unnecessary and potentially harmful. This selective restoration approach reduces computational cost while minimizing the risk of text distortion or blurring by avoiding unnecessary restoration operations.

\paragraph{Tool execution.}
Once the MLLM identifies degradation types that hinder question answering and selects the most severe one whose tool is available, the system invokes the corresponding restoration tool to process the document image. We maintain an available tool list for MLLM selection. The initial tool list contains: \textit{deblurring}, \textit{dewarping}, \textit{deshadowing}, \textit{appearance enhancement}, \textit{demoiréing}, \textit{90° rotation}, \textit{180° rotation}, and \textit{270° rotation}, which respectively target blur, warp, shadow, uneven illumination, moiré pattern, and orientation degradations. We design a standardized tool interface to support future extensions, enabling flexible replacement with more powerful restoration models or integration of new tools for emerging degradation types. Notably, our core contribution is leveraging restoration tools based on task requirements to enhance MLLMs' question answering on degraded documents, rather than improving existing restoration models.

\paragraph{Comparison-based rollback.}
Existing agentic restoration frameworks typically rely on fine-tuned degradation perception models or specialized image quality assessment models to evaluate restoration results and trigger rollback~\cite{zhu2024intelligent,zuo20254kagent}. To avoid additional training and dependence on these specialized models, we design Comparison-Based Rollback. This mechanism provides the MLLM with the current question and the images before and after restoration, allowing it to compare the readability of question-relevant evidence. When restoration distorts or blurs the relevant text and makes the question more difficult to answer, the system triggers rollback (see Figure~\ref{prompt:rollback}). We accordingly determine that the tool cannot effectively address the degradation, disable it from the available tool list, and attempt to address other degradations. When restoration improves the readability of question-relevant evidence and facilitates answering, the system accepts and retains the restored result. As a training-free safeguard, this mechanism prevents harmful restoration results from accumulating across successive operations.

\subsection{Workflow}
\label{sec:workflow}

DocIntent achieves answerability-guided document restoration through a two-phase cyclic mechanism. Given an input real-world document image and question, the system performs Answerability-Guided Restoration in Phase A, followed by Comparison-Based Review in Phase B. This iterative process continues until the MLLM determines that the current image is sufficient to answer the question or that all tools capable of addressing the degradations that hinder answering have been disabled, as illustrated in Figure~\ref{fig:system}.

The two phases coordinate to achieve selective restoration with a quality safeguard. In Phase A, the system determines whether restoration is needed. If the current image suffices for answering, the MLLM generates the answer directly. Otherwise, the system identifies the most severe degradation that hinders answering, invokes the corresponding tool, and proceeds to Phase B. In Phase B, the Comparison-Based Rollback mechanism evaluates the restoration outcome. If textual clarity improves and facilitates question answering, the system accepts the restoration. Otherwise, it triggers a rollback and disables the tool. After either outcome, the system returns to Phase A. Through this cyclic mechanism, DocIntent progressively improves document readability to facilitate question answering.

The design ensures two desirable properties: (1) minimality—only degradations that impair answerability are addressed, avoiding unnecessary processing. (2) safety—every restoration action is validated before being committed, preventing error accumulation across successive operations.

\begin{table*}[t]
  \centering
  \resizebox{\textwidth}{!}{%
  \begin{tabular}{l cc cc cc cc}
    \toprule
    \multicolumn{1}{c}{\multirow{2}{*}{\textbf{Model}}} & \multicolumn{2}{c}{\textbf{WildDocVQA}} & \multicolumn{2}{c}{\textbf{WildChartQA}} & \multicolumn{2}{c}{\textbf{WildTableVQA}} & \multicolumn{2}{c}{\textbf{Average}} \\
    \cmidrule(lr){2-3} \cmidrule(lr){4-5} \cmidrule(lr){6-7} \cmidrule(lr){8-9}
    & ANLS & Consistency & Acc. & Consistency & Acc. & Consistency & Score & Consistency \\
    \midrule
    MiniMonkey-2B~\cite{huang2024mini} & 54.3 & 22.8 & 32.3 & 12.0 & 31.3 & 13.4 & 39.3 & 16.1 \\
    Monkey~\cite{li2024monkey} & 31.0 & 9.9 & 22.4 & 9.8 & 23.0 & 11.7 & 25.5 & 10.5 \\
    TextHarmony~\cite{zhao2024harmonizing} & 37.1 & 16.0 & 21.9 & 10.2 & 14.7 & 6.5 & 24.6 & 10.9 \\
    Llava-Onevision-7B~\cite{li2024llava} & 52.9 & 23.7 & 49.4 & 20.2 & 33.9 & 13.3 & 45.4 & 19.1 \\
    mPLUG-DocOwl2~\cite{hu2025mplug} & 44.2 & 19.2 & 23.3 & 11.6 & 22.1 & 4.8 & 29.9 & 11.9 \\
    MiniCPM-V4.5~\cite{yu2025minicpmv45cookingefficient} & 73.1 & 53.2 & 58.9 & 42.4 & 56.8 & 22.4 & 62.9 & 39.3 \\
    InternVL3.5-8B~\cite{wang2025internvl3_5} & 56.1 & 28.0 & 48.3 & 26.8 & 43.2 & 12.8 & 49.2 & 22.5 \\
    InternVL3.5-38B~\cite{wang2025internvl3_5} & 66.8 & 42.0 & 60.5 & 37.6 & 51.9 & 18.8 & 59.7 & 32.8 \\
    Qwen3.5-9B~\cite{qwen3.5} & 82.5 & 70.8 & 51.9 & 33.2 & 71.1 & 45.2 & 68.5 & 49.7 \\
    Qwen3.5-27B~\cite{qwen3.5} & 83.3 & 72.8 & 53.2 & 43.6 & 76.6 & 49.2 & 71.0 & 55.2 \\
    Qwen3.6-27B~\cite{qwen3.6-27b} & 82.9 & 72.4 & 60.3 & 46.0 & 77.8 & 53.2 & 73.7 & 57.2 \\
    \midrule
    \multicolumn{9}{c}{\textbf{Closed-source MLLMs}}\\
    \midrule
    GPT-4o~\cite{hurst2024gpt} & 63.2 & 41.6 & 31.9 & 23.2 & 59.6 & 30.8 & 51.6 & 31.9 \\
    Qwen3.5-Plus~\cite{qwen3.5} & 84.4 & 77.0 & 62.5 & 52.4 & 81.5 & 59.6 & 76.1 & 63.0 \\
    Claude Sonnet 4.6~\cite{anthropic2025claude4.6sonnet} & 63.4 & 38.4 & 50.0 & 31.6 & 59.1 & 21.6 & 57.5 & 30.5 \\
    Gemini-3.1-Pro~\cite{gemini_3_1_pro_web} & \underline{92.8} & \underline{88.0} & \underline{66.7} & \underline{59.6} & \underline{89.5} & \underline{78.8} & \underline{83.0} & \underline{75.5} \\
    Doubao-2.0-Pro~\cite{seedseed2} & 82.7 & 72.8 & 62.3 & 52.0 & 84.6 & 65.6 & 76.5 & 63.5 \\
    GPT-5.2~\cite{singh2025openai} & 66.1 & 37.6 & 58.3 & 44.8 & 64.2 & 33.6 & 62.9 & 38.7 \\
    \midrule
    \textbf{DocIntent(Qwen3.5-9B)} & 85.4 {\small \textcolor{green!60!black}{+2.9}} & 76.4 {\small \textcolor{green!60!black}{+5.6}} & 53.6 {\small \textcolor{green!60!black}{+1.7}} & 42.0 {\small \textcolor{green!60!black}{+8.8}} & 75.7 {\small \textcolor{green!60!black}{+4.6}} & 48.4 {\small \textcolor{green!60!black}{+3.2}} & 71.6 {\small \textcolor{green!60!black}{+3.1}} & 55.6 {\small \textcolor{green!60!black}{+5.9}} \\
    \textbf{DocIntent(Gemini-3.1-Pro)} & \textbf{94.1} {\small \textcolor{green!60!black}{+1.3}} & \textbf{91.6} {\small \textcolor{green!60!black}{+3.6}} & \textbf{67.7} {\small \textcolor{green!60!black}{+1.0}} & \textbf{63.6} {\small \textcolor{green!60!black}{+4.0}} & \textbf{91.2} {\small \textcolor{green!60!black}{+1.7}} & \textbf{84.8} {\small \textcolor{green!60!black}{+6.0}} & \textbf{84.3} {\small \textcolor{green!60!black}{+1.3}} & \textbf{80.0} {\small \textcolor{green!60!black}{+4.5}} \\
    \bottomrule
  \end{tabular}%
  }
  \caption{Evaluation on the WildDoc~\cite{wang2025wilddoc} benchmark. "ANLS" denotes Average Normalized Levenshtein Similarity. Top results are \textbf{bolded} and second-best \underline{underlined}.}
  \label{tab:baseline_compara}
\end{table*}

\section{Experiments}
\label{sec:experiments}

\subsection{Experimental Settings}
\label{sec:exp_settings}

\paragraph{Implementation details.} All restoration tools are deployed on a single NVIDIA A6000 GPU: DocRes~\cite{zhang2024docres} for deblurring, dewarping, deshadowing, and appearance enhancement, and UniDemoiré~\cite{yang2025unidemoire} for demoiréing (see Appendix~\ref{sec:tool_selection} for tool selection discussions). Open-source MLLMs (e.g., Qwen3.5-9B) run on 4 NVIDIA A6000 GPUs, while closed-source MLLMs are accessed via official APIs. Task-specific prompts for the three WildDoc subtasks are in Appendix~\ref{sec:prompts}. We retain text context (tool decisions and quality assessments) and only the previous image. Context management comparisons and preprocessing and decoding settings are in Appendices~\ref{sec:context_strategy} and~\ref{sec:reproducibility}, respectively.

\paragraph{Datasets.} We adopt WildDoc~\cite{wang2025wilddoc} as our evaluation benchmark. To the best of our knowledge, at the time of our evaluation, WildDoc was the only publicly available benchmark specifically designed for document VQA under real-world capture degradations. It contains three subtasks: WildDocVQA, WildChartQA, and WildTableVQA, covering degradations such as blur, shadow, geometric distortion, moir\'e patterns, uneven illumination, and rotation. Each question corresponds to images captured under four real-world conditions, enabling evaluation of question-answering performance and consistency across degradations.

\paragraph{Metrics.} Following WildDoc~\cite{wang2025wilddoc}, we use Average Normalized Levenshtein Similarity (ANLS) for WildDocVQA, Accuracy (Acc.) for WildChartQA and WildTableVQA, and Consistency (Cons.) for robustness across degradation scenarios. See Appendix~\ref{sec:metrics} for definitions.

\begin{figure*}[t]
  \centering
  \includegraphics[width=\textwidth]{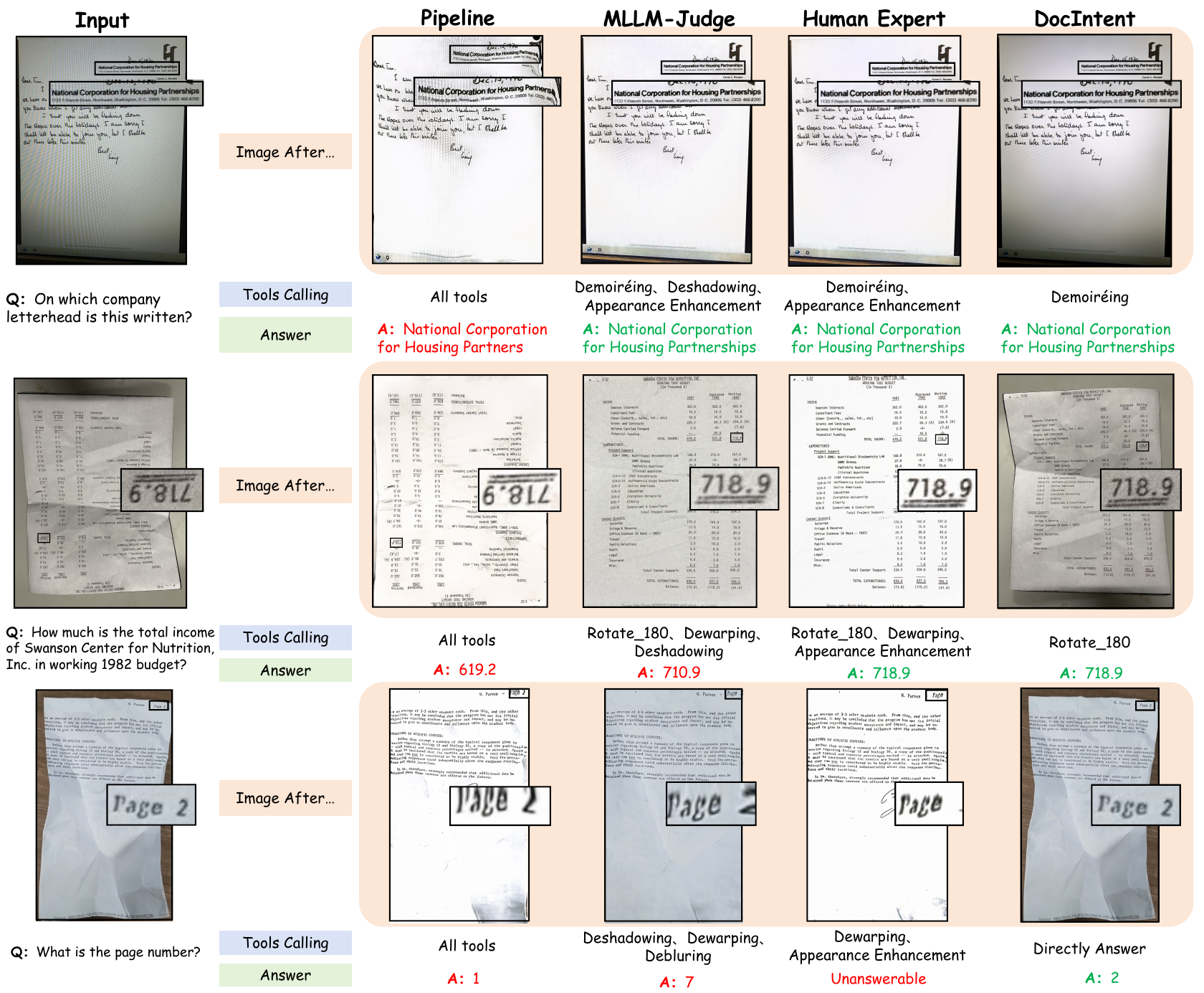}
  \caption{Comparison of restoration strategies. Pipeline blindly applies all tools, MLLM-Judge applies tools to address MLLM-identified degradations, Human Expert uses manually selected tools, and DocIntent performs answerability-guided selective restoration. See Section~\ref{subsec:comparisons} for detailed definitions.}
  \label{fig:complete_compare}
\end{figure*}

\subsection{Main Results}

We evaluate DocIntent on WildDoc with Qwen3.5-9B~\cite{qwen3.5} and Gemini-3.1-Pro~\cite{gemini_3_1_pro_web} as backbone models. Table~\ref{tab:baseline_compara} compares it with a range of open- and closed-source MLLMs.

Among baseline models, we observe a common phenomenon: models' consistency scores are substantially lower than their average score, reflecting their lack of robustness when dealing with different degradation conditions. This score-consistency gap is pervasive across models of varying capabilities. For example, Claude Sonnet 4.6 achieves a 57.5\% average score but only 30.5\% consistency, a 27.0 percentage-point gap. Even the best-performing Gemini-3.1-Pro shows a 7.5 percentage-point gap (83.0\% vs 75.5\%). This widespread phenomenon indicates the fundamental challenge of maintaining stable performance across diverse real-world degradation scenarios.

\begin{table}[t]
  \centering
  \resizebox{\columnwidth}{!}{%
  \begin{tabular}{l l c cc}
    \toprule
    \textbf{Model} & \textbf{Method} & \textbf{Avg. Tools} & \textbf{Avg. Score} & \textbf{Avg. Cons.} \\
    \midrule
    \multirow{8}{*}{Qwen3.5-9B}
    & No restoration & 0.00 & 68.5 & 49.7 \\
    & Pipeline-random  & 5.00 & 63.2 & 45.7 \\
    & Pipeline-fixed  & 5.00 & 63.5 & 47.6 \\
    & MLLM-Judge-random  & 2.78 & 67.7 & 48.8 \\
    & MLLM-Judge-fixed  & 2.78 & 68.5 & 50.7 \\
    & Human Expert & 2.35 & \underline{71.8} & \underline{56.3} \\
    & Human Expert (Question-aware) & 1.55 & \textbf{72.2} & \textbf{57.1} \\
    & \textbf{DocIntent (Ours)} & 1.69 & 71.6 & 55.6 \\
    \midrule
    \multirow{8}{*}{Gemini-3.1-Pro}
    & No restoration & 0.00 & 83.0 & 75.5 \\
    & Pipeline-random  & 5.00 & 79.0 & 68.9 \\
    & Pipeline-fixed  & 5.00 & 79.5 & 69.3 \\
    & MLLM-Judge-random  & 2.47 & 82.9 & 75.3 \\
    & MLLM-Judge-fixed  & 2.47 & 83.1 & 75.2 \\
    & Human Expert & 2.35 & 84.0 & 79.7 \\
    & Human Expert (Question-aware) & 1.55 & \textbf{84.5} & \textbf{80.4} \\
    & \textbf{DocIntent (Ours)} & 1.16 & \underline{84.3} & \underline{80.0} \\
    \bottomrule
  \end{tabular}%
  }
  \caption{Comparison of different restoration strategies on WildDoc benchmark. "Avg." indicates the average results across all three subtasks. Top results are \textbf{bolded} and second-best \underline{underlined}. Complete results for each subtask are provided in Appendix~\ref{sec:complete_results}.}
  \label{tab:strategy_comparison}
\end{table}

DocIntent achieves consistent improvements on both backbone models. When applied to Gemini-3.1-Pro, the average score improves from 83.0\% to 84.3\%. When applied to Qwen3.5-9B, the average score improves from 68.5\% to 71.6\%. Notably, DocIntent demonstrates greater improvements in consistency, with gains of 4.5 points (75.5\% → 80.0\%) on Gemini-3.1-Pro and 5.9 points (49.7\% → 55.6\%) on Qwen3.5-9B. These consistency improvements narrow the score-consistency gap, demonstrating that DocIntent not only improves question-answering performance on degraded documents but also enhances answer stability across diverse degradation scenarios. Figure~\ref{fig:main_case} presents cases where DocIntent enables correct answers on degraded images.

\subsection{Comparisons with Other Strategies}
\label{subsec:comparisons}

Following AgenticIR's evaluation of random tool invocation~\cite{zhu2024intelligent}, we design two restoration baselines, \textbf{Pipeline} and \textbf{MLLM-Judge}. Pipeline applies all image restoration tools sequentially without degradation assessment. MLLM-Judge first uses the MLLM to identify degradations in the image and then invokes the corresponding restoration tools. For both strategies, we test \textbf{random} and \textbf{fixed} tool orders. Random shuffles the tools, whereas fixed follows the sequence of orientation correction, dewarping, demoir\'eing, deblurring, deshadowing, and appearance enhancement. Pipeline excludes orientation correction because it requires MLLM judgment. The \textbf{Human Expert} strategy involves 40 experts with document restoration experience who identify degradations and select tools to optimize overall document readability without access to the corresponding questions. For \textbf{Human Expert (Question-aware)}, experts locate question-relevant evidence and select tools to improve its readability, while skipping restoration when the evidence is already clearly visible. Unlike DocIntent, all comparison strategies answer directly from the preprocessed images without Comparison-Based Rollback.

As shown in Table~\ref{tab:strategy_comparison}, the Pipeline strategy blindly invokes all tools, causing severe information corruption and substantial performance degradation. The MLLM-Judge strategy reduces tool invocations to an average of 2.78 and 2.47 invocations but exhibits performance instability. This is partly due to MLLM misjudgments that may trigger inappropriate tool usage. The fixed sequence and random ordering show only marginal differences, indicating that tool sequencing is not the key limitation. Moreover, even with the comprehensive restoration strategy planned by human experts, restoration tools themselves may introduce text distortion, potentially causing information loss or readability degradation. The absence of a rollback mechanism prevents these errors from being corrected. Both MLLM-Judge and the question-agnostic Human Expert baseline restore all identified degradations, thereby accumulating risks from restoration operations, as illustrated in Figure~\ref{fig:complete_compare}.

The results show that incorporating question awareness helps human experts reduce tool invocations while improving document VQA performance, validating the importance of integrating downstream questions into restoration planning. Without human intervention, DocIntent achieves results close to question-aware human expert preprocessing on both models, demonstrating its potential to automate selective document restoration.

\subsection{Ablation Study}

We analyze the effectiveness of each core component in DocIntent through ablation experiments. As shown in Table~\ref{tab:ablation}, the Prompt-only setting, which enables the Answerability-Guided Prompt while disabling restoration tools and rollback, fails to improve the VQA score and slightly degrades the performance of Qwen3.5-9B. This result shows that more detailed prompting alone cannot explain DocIntent's gains, which primarily come from the selective invocation of restoration tools through Answerability-Guided Restoration (AGRes). With AGRes, the average score and consistency of Gemini-3.1-Pro improve from 83.0\% and 75.5\% to 83.7\% and 78.4\%, respectively. Adding the Comparison-Based Rollback module (CBRb) further improves them to 84.3\% and 80.0\%. On Qwen3.5-9B, AGRes brings modest gains, while the complete DocIntent improves the average score and consistency to 71.6\% and 55.6\%, respectively.

These results demonstrate the effectiveness of both components. For the more capable Gemini-3.1-Pro, AGRes provides effective restoration decisions and brings stable improvements, while CBRb further provides a quality safeguard. For the less capable Qwen3.5-9B, AGRes brings limited gains, making CBRb more critical. CBRb promptly identifies and rolls back inappropriate restorations through quality comparison (Figure~\ref{fig:case_rollback}), bringing greater performance gains to weaker models. These results validate the complete system design: answerability-guided restoration provides selective processing capability, the rollback mechanism provides a quality safeguard, and their synergy achieves stable improvements across models of different capability levels.

\begin{table}[t]
  \centering
  \resizebox{\columnwidth}{!}{%
  \begin{tabular}{l ccc cc}
    \toprule
    \textbf{Model} & \textbf{AG Prompt} & \textbf{AGRes.} & \textbf{CBRb.} & \textbf{Avg. Score} & \textbf{Avg. Cons.} \\
    \midrule
    \multirow{4}{*}{Qwen3.5-9B}
    & \textcolor{red!85}{\ding{55}} & \textcolor{red!85}{\ding{55}} & \textcolor{red!85}{\ding{55}} & 68.5 & 49.7 \\
    & \textcolor{green!85}{\ding{51}} & \textcolor{red!85}{\ding{55}} & \textcolor{red!85}{\ding{55}} & 67.8 & 49.2 \\
    & \textcolor{green!85}{\ding{51}} & \textcolor{green!85}{\ding{51}} & \textcolor{red!85}{\ding{55}} & \underline{69.2} & \underline{50.5} \\
    & \textcolor{green!85}{\ding{51}} & \textcolor{green!85}{\ding{51}} & \textcolor{green!85}{\ding{51}} & \textbf{71.6} & \textbf{55.6} \\
    \midrule
    \multirow{4}{*}{Gemini-3.1-Pro}
    & \textcolor{red!85}{\ding{55}} & \textcolor{red!85}{\ding{55}} & \textcolor{red!85}{\ding{55}} & 83.0 & 75.5 \\
    & \textcolor{green!85}{\ding{51}} & \textcolor{red!85}{\ding{55}} & \textcolor{red!85}{\ding{55}} & 82.9 & 75.7 \\
    & \textcolor{green!85}{\ding{51}} & \textcolor{green!85}{\ding{51}} & \textcolor{red!85}{\ding{55}} & \underline{83.7} & \underline{78.4} \\
    & \textcolor{green!85}{\ding{51}} & \textcolor{green!85}{\ding{51}} & \textcolor{green!85}{\ding{51}} & \textbf{84.3} & \textbf{80.0} \\
    \bottomrule
  \end{tabular}%
  }
  \caption{Ablation study on DocIntent components on WildDoc benchmark. "AG Prompt" denotes answerability-guided prompting. The Prompt-only setting enables AG Prompt while disabling AGRes and CBRb. "Avg." indicates the average results across all three subtasks. Complete results for each subtask are provided in Appendix~\ref{sec:complete_results}.}
  \label{tab:ablation}
\end{table}

\section{Conclusion}
\label{sec:conclusion}

In this paper, we demonstrate that beyond the quality of restoration tools, the critical factor for document restoration in VQA is the intelligence of the orchestration strategy: determining when to restore, which tool to use, and whether to accept the results. Answerability-guided selective restoration is essential for performance improvement. Based on these insights, we present DocIntent, a training-free document restoration framework that achieves selective restoration with a quality safeguard through a two-phase cyclic mechanism. Extensive experiments demonstrate that our method improves models' question-answering performance on degraded documents through efficient tool utilization.

\section*{Limitations} While our approach efficiently utilizes restoration tools to assist MLLMs in document question answering on degraded images, the system's performance is fundamentally limited by the capabilities of the underlying restoration tools. Additionally, as a training-free agentic framework, the system's judgment capability depends on the intrinsic abilities of the underlying MLLM.

\section*{Ethical Considerations} This research poses no significant ethical risks. All experiments are conducted on publicly available open-source datasets, without involving any personally identifiable information or sensitive content. This work only restores degraded documents and does not involve content tampering or forgery, ensuring no additional ethical concerns.

\bibliography{custom}

\clearpage
\appendix
\section*{Appendix} 

\section{Evaluation Metrics}
\label{sec:metrics}

We provide detailed definitions of the evaluation metrics used in our experiments. 

The accuracy metric quantifies the proportion of questions where the predicted answer precisely corresponds with any of the designated target answers for that question. 

For the Average Normalized Levenshtein Similarity (ANLS), we follow previous works~\cite{mathew2021docvqa}, which is defined as follows:

\begin{equation}
\text{ANLS} = \frac{1}{N} \sum_{i=1}^{N} \left( \max\nolimits_{j} s(a_{ij}, o_{q_i}) \right)
\end{equation}

where $s(a_{ij}, o_{q_i})$ is defined as $1 - NL(a_{ij}, o_{q_i})$ when $NL(a_{ij}, o_{q_i})$, the normalized Levenshtein distance, is less than a predefined threshold $\tau$; otherwise 0. We set the threshold $\tau = 0.5$, as previous works do~\cite{mathew2021docvqa}.

Following WildDoc~\cite{wang2025wilddoc}, we also report the consistency score, a robustness metric designed to assess the resilience of models when handling the same question across images captured under various conditions. This metric calculates the document-level accuracy. A model's response is considered accurate only if it correctly answers the question in all four distinct scenarios presented.

\section{Algorithm Workflow}
\label{sec:algorithm_workflow}

Algorithm~\ref{alg:workflow} describes the complete inference workflow of DocIntent. The functions Analyze, SelectTool, Compare, and Answer are all implemented by the MLLM. In the Analyze function, the MLLM analyzes the current image and question to determine whether it can be directly answered. In the SelectTool function, the MLLM identifies degradation types that hinder answering and selects the tool corresponding to the most severe degradation from the available tool list. The Execute function invokes the actual restoration tools. In the Compare function, the MLLM compares the images before and after restoration with respect to the question to determine whether the restoration improves the readability of question-relevant evidence. In the Answer function, the MLLM generates the answer to the question based on the final image.

We set a maximum iteration count $T$ for the inference workflow to prevent the system from falling into excessive repetitive tool invocations in extreme cases where the comparison-based rollback mechanism fails. In all experiments, we set $T=10$, providing sufficient restoration redundancy for the system.

\begin{table*}[t]
  \centering
  \resizebox{\linewidth}{!}{%
  \begin{tabular}{l cc cc cc cc}
    \toprule
    \multicolumn{1}{c}{\multirow{2}{*}{\textbf{Model}}} & \multicolumn{2}{c}{\textbf{WildDocVQA}} & \multicolumn{2}{c}{\textbf{WildChartQA}} & \multicolumn{2}{c}{\textbf{WildTableVQA}} & \multicolumn{2}{c}{\textbf{Average}} \\
    \cmidrule(lr){2-3} \cmidrule(lr){4-5} \cmidrule(lr){6-7} \cmidrule(lr){8-9}
    & ANLS & Consistency & Acc. & Consistency & Acc. & Consistency & Score & Consistency \\
    \midrule
    Qwen3.5-27B~\cite{qwen3.5} & 83.3 & 72.8 & 53.2 & 43.6 & 76.6 & 49.2 & 71.0 & 55.2 \\
    \textbf{w/ DocIntent} & \textbf{84.1} {\small \textcolor{green!60!black}{+0.8}} & \textbf{74.4} {\small \textcolor{green!60!black}{+1.6}} & \textbf{54.9} {\small \textcolor{green!60!black}{+1.7}} & \textbf{50.8} {\small \textcolor{green!60!black}{+7.2}} & \textbf{86.3} {\small \textcolor{green!60!black}{+9.7}} & \textbf{58.8} {\small \textcolor{green!60!black}{+9.6}} & \textbf{75.1} {\small \textcolor{green!60!black}{+4.1}} & \textbf{61.3} {\small \textcolor{green!60!black}{+6.1}} \\
    \midrule
    GPT-4o~\cite{hurst2024gpt} & 63.2 & 41.6 & 31.9 & 23.2 & 59.6 & 30.8 & 51.6 & 31.9 \\
    \textbf{w/ DocIntent} & \textbf{73.7} {\small \textcolor{green!60!black}{+10.5}} & \textbf{58.4} {\small \textcolor{green!60!black}{+16.8}} & \textbf{35.8} {\small \textcolor{green!60!black}{+3.9}} & \textbf{34.4} {\small \textcolor{green!60!black}{+11.2}} & \textbf{64.2} {\small \textcolor{green!60!black}{+4.6}} & \textbf{38.4} {\small \textcolor{green!60!black}{+7.6}} & \textbf{57.9} {\small \textcolor{green!60!black}{+6.3}} & \textbf{43.7} {\small \textcolor{green!60!black}{+11.8}} \\
    \midrule
    Qwen3.5-Plus~\cite{qwen3.5} & 84.4 & 77.0 & 62.5 & 52.4 & 81.5 & 59.6 & 76.1 & 63.0 \\
    \textbf{w/ DocIntent} & \textbf{91.0} {\small \textcolor{green!60!black}{+6.6}} & \textbf{82.4} {\small \textcolor{green!60!black}{+5.4}} & \textbf{67.9} {\small \textcolor{green!60!black}{+5.4}} & \textbf{59.6} {\small \textcolor{green!60!black}{+7.2}} & \textbf{86.1} {\small \textcolor{green!60!black}{+4.6}} & \textbf{67.6} {\small \textcolor{green!60!black}{+8.0}} & \textbf{81.7} {\small \textcolor{green!60!black}{+5.6}} & \textbf{69.9} {\small \textcolor{green!60!black}{+6.9}} \\
    \midrule
    Claude Sonnet 4.6~\cite{anthropic2025claude4.6sonnet} & 63.4 & 38.4 & 50.0 & 31.6 & 59.1 & 21.6 & 57.5 & 30.5 \\
    \textbf{w/ DocIntent} & \textbf{71.9} {\small \textcolor{green!60!black}{+8.5}} & \textbf{49.6} {\small \textcolor{green!60!black}{+11.2}} & \textbf{57.0} {\small \textcolor{green!60!black}{+7.0}} & \textbf{39.2} {\small \textcolor{green!60!black}{+7.6}} & \textbf{70.3} {\small \textcolor{green!60!black}{+11.2}} & \textbf{36.4} {\small \textcolor{green!60!black}{+14.8}} & \textbf{66.4} {\small \textcolor{green!60!black}{+8.9}} & \textbf{41.7} {\small \textcolor{green!60!black}{+11.2}} \\
    \midrule
    Doubao-2.0-Pro~\cite{seedseed2} & 82.7 & 72.8 & 62.3 & 52.0 & 84.6 & 65.6 & 76.5 & 63.5 \\
    \textbf{w/ DocIntent} & \textbf{83.0} {\small \textcolor{green!60!black}{+0.3}} & \textbf{74.8} {\small \textcolor{green!60!black}{+2.0}} & \textbf{67.4} {\small \textcolor{green!60!black}{+5.1}} & \textbf{58.0} {\small \textcolor{green!60!black}{+6.0}} & \textbf{84.9} {\small \textcolor{green!60!black}{+0.3}} & \textbf{66.4} {\small \textcolor{green!60!black}{+0.8}} & \textbf{78.4} {\small \textcolor{green!60!black}{+1.9}} & \textbf{66.4} {\small \textcolor{green!60!black}{+2.9}} \\
    \midrule
    GPT-5.2~\cite{singh2025openai} & 66.1 & 37.6 & 58.3 & 44.8 & 64.2 & 33.6 & 62.9 & 38.7 \\
    \textbf{w/ DocIntent} & \textbf{76.7} {\small \textcolor{green!60!black}{+10.6}} & \textbf{52.8} {\small \textcolor{green!60!black}{+15.2}} & \textbf{62.5} {\small \textcolor{green!60!black}{+4.2}} & \textbf{48.8} {\small \textcolor{green!60!black}{+4.0}} & \textbf{69.4} {\small \textcolor{green!60!black}{+5.2}} & \textbf{41.6} {\small \textcolor{green!60!black}{+8.0}} & \textbf{69.5} {\small \textcolor{green!60!black}{+6.6}} & \textbf{47.7} {\small \textcolor{green!60!black}{+9.0}} \\
    \bottomrule
  \end{tabular}%
  }
  \caption{Performance of additional MLLMs w/ and w/o DocIntent on WildDoc benchmark~\cite{wang2025wilddoc}. "ANLS" denotes Average Normalized Levenshtein Similarity.}
  \label{tab:adition}
\end{table*}

\begin{algorithm}[t]
\caption{DocIntent Workflow}
\label{alg:workflow}
\begin{algorithmic}[1]
\REQUIRE Initial document image $I_0$, Question $Q$, Maximum iterations $T$
\ENSURE Answer $A$

\STATE $tools \leftarrow \text{ALL\_TOOLS}$

\FOR{$t = 0$ to $T-1$}

    \STATE $can\_answer \leftarrow \text{Analyze}(I_{t}, Q)$
    \IF{$can\_answer$}
        \RETURN $\text{Answer}(I_{t}, Q)$
    \ENDIF

    \STATE $tool \leftarrow \text{SelectTool}(I_{t}, Q, tools)$
    \IF{$tool$ is $None$}
        \RETURN $\text{Answer}(I_{t}, Q)$
    \ENDIF

    \STATE $I_{t+1} \leftarrow \text{Execute}(tool, I_{t})$

    \STATE $is\_improved \leftarrow \text{Compare}(I_{t}, I_{t+1}, Q)$
    \IF{\textbf{not} $is\_improved$}
        \STATE $I_{t+1} \leftarrow I_{t}$
        \STATE $tools \leftarrow tools \setminus \{tool\}$
    \ENDIF

\ENDFOR

\RETURN $\text{Answer}(I_{T}, Q)$

\end{algorithmic}
\end{algorithm}

\section{Additional Experiments}
\label{sec:Additional_Experiments}

To evaluate DocIntent across models with different capability levels, we conduct additional experiments on more MLLMs. As shown in Table~\ref{tab:adition}, DocIntent achieves performance improvements across all tested models. These additional experiments cover the open-source model Qwen3.5-27B~\cite{qwen3.5} and closed-source models including GPT-4o~\cite{hurst2024gpt}, Qwen3.5-Plus~\cite{qwen3.5}, Claude Sonnet 4.6~\cite{anthropic2025claude4.6sonnet}, Doubao-2.0-Pro~\cite{seedseed2}, and GPT-5.2~\cite{singh2025openai}, showing consistent improvements on the evaluated MLLMs.

\begin{table*}[t]
  \centering
  \resizebox{\textwidth}{!}{%
  \begin{tabular}{l l ccc ccc ccc ccc}
    \toprule
    \multicolumn{1}{c}{\multirow{2}{*}{\textbf{Model}}} & \multicolumn{1}{c}{\multirow{2}{*}{\textbf{Context Strategy}}} & \multicolumn{3}{c}{\textbf{WildDocVQA}} & \multicolumn{3}{c}{\textbf{WildChartQA}} & \multicolumn{3}{c}{\textbf{WildTableVQA}} & \multicolumn{3}{c}{\textbf{Average}} \\
    \cmidrule(lr){3-5} \cmidrule(lr){6-8} \cmidrule(lr){9-11} \cmidrule(lr){12-14}
    & & ANLS & Consistency & Tokens & Acc. & Consistency & Tokens & Acc. & Consistency & Tokens & Score & Consistency & Tokens \\
    \midrule
    \multirow{3}{*}{Qwen3.5-9B}
    & No Context & 84.6 & 75.6 & 15880 & 40.7 & 32.8 & 29792 & 71.5 & 43.4 & 25826 & 65.6 & 50.6 & 23833 \\
    & Full Context & \textbf{86.1} & \underline{76.0} & 27967 & \textbf{54.9} & \underline{41.6} & 30735 & \underline{73.5} & \underline{46.0} & 29615 & \underline{71.5} & \underline{54.5} & 29439 \\
    & \textbf{Text-only (Ours)} & \underline{85.4} & \textbf{76.4} & 17967 & \underline{53.6} & \textbf{42.0} & 19611 & \textbf{75.7} & \textbf{48.4} & 19518 & \textbf{71.6} & \textbf{55.6} & 19032 \\
    \midrule
    \multirow{3}{*}{Gemini-3.1-Pro}
    & No Context & 92.5 & 88.0 & 4904 & 61.6 & 56.4 & 5322 & 90.7 & 81.2 & 4322 & 81.6 & 75.2 & 4849 \\
    & Full Context & \underline{94.0} & \textbf{92.0} & 7352 & \underline{66.3} & \underline{62.8} & 7625 & \textbf{92.4} & \textbf{85.7} & 7147 & \underline{84.2} & \textbf{80.2} & 7375 \\
    & \textbf{Text-only (Ours)} & \textbf{94.1} & \underline{91.6} & 5817 & \textbf{67.7} & \textbf{63.6} & 6791 & \underline{91.2} & \underline{84.8} & 5711 & \textbf{84.3} & \underline{80.0} & 6106 \\
    \bottomrule
  \end{tabular}%
  }
  \caption{Comparison of different context management strategies. "ANLS" denotes Average Normalized Levenshtein Similarity. Within each model, top results are \textbf{bolded} and second-best \underline{underlined}.}
  \label{tab:context_strategy}
\end{table*}

\begin{table}[t]
  \centering
  \resizebox{\columnwidth}{!}{%
  \begin{tabular}{l l c c c c}
    \toprule
    \multicolumn{1}{c}{\textbf{Model}} & \textbf{Category} & \textbf{WildDocVQA} & \textbf{WildChartQA} & \textbf{WildTableVQA} & \textbf{Average} \\
    & & ANLS & Acc. & Acc. & Score \\
    \midrule
    \multirow{3}{*}{Qwen3.5-9B}
    & \textbf{Baseline} & \textbf{82.5} & \textbf{51.9} & \textbf{71.1} & \textbf{68.5} \\
    & Confident & 84.6 {\small \textcolor{green!60!black}{+2.1}} & 54.4 {\small \textcolor{green!60!black}{+2.5}} & 83.7 {\small \textcolor{green!60!black}{+12.6}} & 74.2 {\small \textcolor{green!60!black}{+5.7}} \\
    & Forced & 78.6 {\small \textcolor{red!60!black}{-3.9}} & 48.8 {\small \textcolor{red!60!black}{-3.1}} & 57.6 {\small \textcolor{red!60!black}{-13.5}} & 61.7 {\small \textcolor{red!60!black}{-6.8}} \\
    \midrule
    \multirow{3}{*}{Gemini-3.1-Pro}
    & \textbf{Baseline} & \textbf{92.8} & \textbf{66.7} & \textbf{89.5} & \textbf{83.0} \\
    & Confident & 95.6 {\small \textcolor{green!60!black}{+2.8}} & 70.9 {\small \textcolor{green!60!black}{+4.2}} & 91.1 {\small \textcolor{green!60!black}{+1.6}} & 85.9 {\small \textcolor{green!60!black}{+2.9}} \\
    & Forced & 90.3 {\small \textcolor{red!60!black}{-2.5}} & 60.7 {\small \textcolor{red!60!black}{-6.0}} & 87.6 {\small \textcolor{red!60!black}{-1.9}} & 79.5 {\small \textcolor{red!60!black}{-3.5}} \\
    \midrule
    \multirow{3}{*}{Qwen3.5-27B}
    & \textbf{Baseline} & \textbf{83.3} & \textbf{53.2} & \textbf{76.6} & \textbf{71.0} \\
    & Confident & 90.0 {\small \textcolor{green!60!black}{+6.7}} & 61.3 {\small \textcolor{green!60!black}{+8.1}} & 83.2 {\small \textcolor{green!60!black}{+6.6}} & 78.2 {\small \textcolor{green!60!black}{+7.2}} \\
    & Forced & 72.5 {\small \textcolor{red!60!black}{-10.8}} & 41.6 {\small \textcolor{red!60!black}{-11.6}} & 65.7 {\small \textcolor{red!60!black}{-10.9}} & 59.9 {\small \textcolor{red!60!black}{-11.1}} \\
    \midrule
    \multirow{3}{*}{GPT-4o}
    & \textbf{Baseline} & \textbf{63.2} & \textbf{31.9} & \textbf{59.6} & \textbf{51.6} \\
    & Confident & 86.3 {\small \textcolor{green!60!black}{+23.1}} & 40.0 {\small \textcolor{green!60!black}{+8.1}} & 79.7 {\small \textcolor{green!60!black}{+20.1}} & 68.7 {\small \textcolor{green!60!black}{+17.1}} \\
    & Forced & 43.4 {\small \textcolor{red!60!black}{-19.8}} & 26.7 {\small \textcolor{red!60!black}{-5.2}} & 45.4 {\small \textcolor{red!60!black}{-14.2}} & 38.5 {\small \textcolor{red!60!black}{-13.1}} \\
    \midrule
    \multirow{3}{*}{Qwen3.5-Plus}
    & \textbf{Baseline} & \textbf{84.4} & \textbf{62.5} & \textbf{81.5} & \textbf{76.1} \\
    & Confident & 95.5 {\small \textcolor{green!60!black}{+11.1}} & 65.5 {\small \textcolor{green!60!black}{+3.0}} & 90.9 {\small \textcolor{green!60!black}{+9.4}} & 84.0 {\small \textcolor{green!60!black}{+7.9}} \\
    & Forced & 73.4 {\small \textcolor{red!60!black}{-11.0}} & 60.4 {\small \textcolor{red!60!black}{-2.1}} & 74.3 {\small \textcolor{red!60!black}{-7.2}} & 69.4 {\small \textcolor{red!60!black}{-6.7}} \\
    \midrule
    \multirow{3}{*}{Claude Sonnet 4.6}
    & \textbf{Baseline} & \textbf{63.4} & \textbf{50.0} & \textbf{59.1} & \textbf{57.5} \\
    & Confident & 85.8 {\small \textcolor{green!60!black}{+22.4}} & 58.7 {\small \textcolor{green!60!black}{+8.7}} & 77.1 {\small \textcolor{green!60!black}{+18.0}} & 73.9 {\small \textcolor{green!60!black}{+16.4}} \\
    & Forced & 56.7 {\small \textcolor{red!60!black}{-6.7}} & 37.7 {\small \textcolor{red!60!black}{-12.3}} & 44.7 {\small \textcolor{red!60!black}{-14.4}} & 46.4 {\small \textcolor{red!60!black}{-11.1}} \\
    \midrule
    \multirow{3}{*}{Doubao-2.0-Pro}
    & \textbf{Baseline} & \textbf{82.7} & \textbf{62.3} & \textbf{84.6} & \textbf{76.5} \\
    & Confident & 87.0 {\small \textcolor{green!60!black}{+4.3}} & 64.5 {\small \textcolor{green!60!black}{+2.2}} & 87.5 {\small \textcolor{green!60!black}{+2.9}} & 79.7 {\small \textcolor{green!60!black}{+3.2}} \\
    & Forced & 69.5 {\small \textcolor{red!60!black}{-13.2}} & 56.1 {\small \textcolor{red!60!black}{-6.2}} & 67.0 {\small \textcolor{red!60!black}{-17.6}} & 64.2 {\small \textcolor{red!60!black}{-12.3}} \\
    \midrule
    \multirow{3}{*}{GPT-5.2}
    & \textbf{Baseline} & \textbf{66.1} & \textbf{58.3} & \textbf{64.2} & \textbf{62.9} \\
    & Confident & 87.3 {\small \textcolor{green!60!black}{+21.2}} & 64.3 {\small \textcolor{green!60!black}{+6.0}} & 78.7 {\small \textcolor{green!60!black}{+14.5}} & 76.8 {\small \textcolor{green!60!black}{+13.9}} \\
    & Forced & 41.9 {\small \textcolor{red!60!black}{-24.2}} & 50.2 {\small \textcolor{red!60!black}{-8.1}} & 50.4 {\small \textcolor{red!60!black}{-13.8}} & 47.5 {\small \textcolor{red!60!black}{-15.4}} \\
    \bottomrule
  \end{tabular}%
  }
  \caption{Analysis of tool invocation decisions where all categories employ direct answering w/o DocIntent. "ANLS" denotes Average Normalized Levenshtein Similarity.}
  \label{tab:tool_decision}
\end{table}

\section{Analysis of Tool Invocation Decisions}

Different MLLMs vary in their sensitivity to degradation types; tool invocation decisions should therefore be analyzed in the context of the particular model and downstream task. Table~\ref{tab:tool_decision} uses direct answering for all categories and partitions samples into the Confident subset, which the model judges answerable directly, and the Forced subset, which the model judges require tool assistance. The Confident subset usually scores higher than the overall baseline, whereas the Forced subset scores lower, indicating that answerability judgments distinguish samples with different levels of direct-answering difficulty.

To further assess the downstream utility of tool invocation, Table~\ref{tab:forced_subset} compares Direct Answer and DocIntent on each model's Forced subset. DocIntent obtains higher scores for all models and subtasks, showing that Answerability-Guided Restoration provides consistent aggregate gains on samples that are more difficult to answer directly. Together, the two tables show that DocIntent improves downstream VQA performance through selective restoration on samples judged to require tool assistance.

\begin{table}[t]
  \centering
  \resizebox{\columnwidth}{!}{%
  \begin{tabular}{llcccc}
    \toprule
    \multicolumn{1}{c}{\multirow{2}{*}{\textbf{Model}}} & \multicolumn{1}{c}{\multirow{2}{*}{\textbf{Method}}} & \multicolumn{1}{c}{\textbf{WildDocVQA}} & \multicolumn{1}{c}{\textbf{WildChartQA}} & \multicolumn{1}{c}{\textbf{WildTableVQA}} & \multicolumn{1}{c}{\textbf{Average}} \\
    \cmidrule(lr){3-3} \cmidrule(lr){4-4} \cmidrule(lr){5-5}
    & & ANLS & Acc. & Acc. & Score \\
    \midrule
    \multirow{2}{*}{Qwen3.5-9B}
    & Direct Answer & 78.6 & 48.8 & 57.6 & 61.7 \\
    & \textbf{DocIntent} & \textbf{86.9} {\small \textcolor{green!60!black}{+8.3}} & \textbf{52.6} {\small \textcolor{green!60!black}{+3.8}} & \textbf{67.1} {\small \textcolor{green!60!black}{+9.5}} & \textbf{68.9} {\small \textcolor{green!60!black}{+7.2}} \\
    \midrule
    \multirow{2}{*}{Gemini-3.1-Pro}
    & Direct Answer & 90.3 & 60.7 & 87.6 & 79.5 \\
    & \textbf{DocIntent} & \textbf{92.8} {\small \textcolor{green!60!black}{+2.5}} & \textbf{63.1} {\small \textcolor{green!60!black}{+2.4}} & \textbf{91.3} {\small \textcolor{green!60!black}{+3.7}} & \textbf{82.4} {\small \textcolor{green!60!black}{+2.9}} \\
    \midrule
    \multirow{2}{*}{Qwen3.5-27B}
    & Direct Answer & 72.5 & 41.6 & 65.7 & 59.9 \\
    & \textbf{DocIntent} & \textbf{74.6} {\small \textcolor{green!60!black}{+2.1}} & \textbf{45.7} {\small \textcolor{green!60!black}{+4.1}} & \textbf{91.4} {\small \textcolor{green!60!black}{+25.7}} & \textbf{70.6} {\small \textcolor{green!60!black}{+10.7}} \\
    \midrule
    \multirow{2}{*}{GPT-4o}
    & Direct Answer & 43.4 & 26.7 & 45.4 & 38.5 \\
    & \textbf{DocIntent} & \textbf{62.9} {\small \textcolor{green!60!black}{+19.5}} & \textbf{33.1} {\small \textcolor{green!60!black}{+6.4}} & \textbf{53.3} {\small \textcolor{green!60!black}{+7.9}} & \textbf{49.8} {\small \textcolor{green!60!black}{+11.3}} \\
    \midrule
    \multirow{2}{*}{Qwen3.5-Plus}
    & Direct Answer & 73.4 & 60.4 & 74.3 & 69.4 \\
    & \textbf{DocIntent} & \textbf{86.5} {\small \textcolor{green!60!black}{+13.1}} & \textbf{69.6} {\small \textcolor{green!60!black}{+9.2}} & \textbf{82.4} {\small \textcolor{green!60!black}{+8.1}} & \textbf{79.5} {\small \textcolor{green!60!black}{+10.1}} \\
    \midrule
    \multirow{2}{*}{Claude Sonnet 4.6}
    & Direct Answer & 56.7 & 37.7 & 44.7 & 46.4 \\
    & \textbf{DocIntent} & \textbf{67.7} {\small \textcolor{green!60!black}{+11.0}} & \textbf{54.6} {\small \textcolor{green!60!black}{+16.9}} & \textbf{64.9} {\small \textcolor{green!60!black}{+20.2}} & \textbf{62.4} {\small \textcolor{green!60!black}{+16.0}} \\
    \midrule
    \multirow{2}{*}{Doubao-2.0-Pro}
    & Direct Answer & 69.5 & 56.1 & 67.0 & 64.2 \\
    & \textbf{DocIntent} & \textbf{70.7} {\small \textcolor{green!60!black}{+1.2}} & \textbf{75.6} {\small \textcolor{green!60!black}{+19.5}} & \textbf{69.1} {\small \textcolor{green!60!black}{+2.1}} & \textbf{71.8} {\small \textcolor{green!60!black}{+7.6}} \\
    \midrule
    \multirow{2}{*}{GPT-5.2}
    & Direct Answer & 41.9 & 50.2 & 50.4 & 47.5 \\
    & \textbf{DocIntent} & \textbf{64.6} {\small \textcolor{green!60!black}{+22.7}} & \textbf{60.1} {\small \textcolor{green!60!black}{+9.9}} & \textbf{60.5} {\small \textcolor{green!60!black}{+10.1}} & \textbf{61.7} {\small \textcolor{green!60!black}{+14.2}} \\
    \bottomrule
  \end{tabular}
  }
  \caption{Comparison on the Forced subset, where DocIntent decides to invoke restoration tools. "ANLS" denotes Average Normalized Levenshtein Similarity.}
  \label{tab:forced_subset}
\end{table}

\begin{table*}[t]
  \centering
  \resizebox{\textwidth}{!}{%
  \begin{tabular}{l l c cc cc cc cc}
    \toprule
    \multicolumn{1}{c}{\multirow{2}{*}{\textbf{Model}}} & \multicolumn{1}{c}{\multirow{2}{*}{\textbf{Method}}} & \multirow{2}{*}{\textbf{Avg. Tools}} & \multicolumn{2}{c}{\textbf{WildDocVQA}} & \multicolumn{2}{c}{\textbf{WildChartQA}} & \multicolumn{2}{c}{\textbf{WildTableVQA}} & \multicolumn{2}{c}{\textbf{Average}} \\
    \cmidrule(lr){4-5} \cmidrule(lr){6-7} \cmidrule(lr){8-9} \cmidrule(lr){10-11}
    & & & ANLS & Consistency & Acc. & Consistency & Acc. & Consistency & Score & Consistency \\
    \midrule
    \multirow{8}{*}{Qwen3.5-9B}
    & No restoration & 0.00 & 82.5 & 70.8 & 51.9 & 33.2 & 71.1 & 45.2 & 68.5 & 49.7 \\
    & Pipeline-random  & 5.00 & 78.2 & 62.4 & 51.4 & 32.8 & 60.0 & 42.0 & 63.2 & 45.7 \\
    & Pipeline-fixed  & 5.00 & 80.6 & 69.6 & 51.0 & 31.6 & 58.9 & 41.6 & 63.5 & 47.6 \\
    & MLLM-Judge-random  & 2.78 & 82.9 & 68.8 & 50.1 & 32.0 & 70.1 & 45.6 & 67.7 & 48.8 \\
    & MLLM-Judge-fixed  & 2.78 & 82.7 & 70.6 & 51.7 & 34.8 & 71.2 & 46.8 & 68.5 & 50.7 \\
    & Human Expert & 2.35 & \underline{85.7} & 75.6 & \textbf{53.6} & \underline{43.2} & \underline{76.2} & \underline{50.0} & \underline{71.8} & \underline{56.3} \\
    & Human Expert (Question-aware) & 1.55 & \textbf{86.3} & \textbf{76.8} & \underline{53.5} & \textbf{43.6} & \textbf{76.9} & \textbf{50.8} & \textbf{72.2} & \textbf{57.1} \\
    & \textbf{DocIntent (Ours)} & 1.69 & 85.4 & \underline{76.4} & \textbf{53.6} & 42.0 & 75.7 & 48.4 & 71.6 & 55.6 \\
    \midrule
    \multirow{8}{*}{Gemini-3.1-Pro}
    & No restoration & 0.00 & 92.8 & 88.0 & 66.7 & 59.6 & 89.5 & 78.8 & 83.0 & 75.5 \\
    & Pipeline-random  & 5.00 & 89.1 & 75.2 & 61.4 & 54.4 & 86.4 & 77.2 & 79.0 & 68.9 \\
    & Pipeline-fixed  & 5.00 & 89.7 & 77.2 & 61.7 & 54.0 & 87.0 & 76.8 & 79.5 & 69.3 \\
    & MLLM-Judge-random  & 2.47 & 92.7 & 87.2 & 65.4 & 57.2 & 90.6 & 81.6 & 82.9 & 75.3 \\
    & MLLM-Judge-fixed  & 2.47 & 92.2 & 86.4 & 66.2 & 57.6 & 90.9 & 81.6 & 83.1 & 75.2 \\
    & Human Expert & 2.35 & 93.5 & 89.6 & 67.4 & \textbf{65.2} & 91.0 & \underline{84.4} & 84.0 & 79.7 \\
    & Human Expert (Question-aware) & 1.55 & \underline{94.0} & \underline{91.3} & \textbf{68.0} & \textbf{65.2} & \textbf{91.4} & \textbf{84.8} & \textbf{84.5} & \textbf{80.4} \\
    & \textbf{DocIntent (Ours)} & 1.16 & \textbf{94.1} & \textbf{91.6} & \underline{67.7} & \underline{63.6} & \underline{91.2} & \textbf{84.8} & \underline{84.3} & \underline{80.0} \\
    \bottomrule
  \end{tabular}%
  }
  \caption{Complete results of restoration strategy comparison on WildDoc benchmark. "ANLS" denotes Average Normalized Levenshtein Similarity. "Avg. Tools" denotes the average number of tool invocations per sample. Top results are \textbf{bolded} and second-best \underline{underlined}.}
  \label{tab:strategy_comparison_full}
\end{table*}

\begin{table*}[t]
  \centering
  \resizebox{\textwidth}{!}{%
  \begin{tabular}{l ccc cc cc cc cc}
    \toprule
    \multicolumn{1}{c}{\multirow{2}{*}{\textbf{Model}}} & \multirow{2}{*}{\textbf{AG Prompt}} & \multirow{2}{*}{\textbf{AGRes.}} & \multirow{2}{*}{\textbf{CBRb.}} & \multicolumn{2}{c}{\textbf{WildDocVQA}} & \multicolumn{2}{c}{\textbf{WildChartQA}} & \multicolumn{2}{c}{\textbf{WildTableVQA}} & \multicolumn{2}{c}{\textbf{Average}} \\
    \cmidrule(lr){5-6} \cmidrule(lr){7-8} \cmidrule(lr){9-10} \cmidrule(lr){11-12}
    & & & & ANLS & Consistency & Acc. & Consistency & Acc. & Consistency & Score & Consistency \\
    \midrule
    \multirow{4}{*}{Qwen3.5-9B}
    & \textcolor{red!85}{\ding{55}} & \textcolor{red!85}{\ding{55}} & \textcolor{red!85}{\ding{55}} & 82.5 & \underline{70.8} & 51.9 & 33.2 & 71.1 & 45.2 & 68.5 & 49.7 \\
    & \textcolor{green!85}{\ding{51}} & \textcolor{red!85}{\ding{55}} & \textcolor{red!85}{\ding{55}} & 82.0 & 70.3 & 51.0 & 33.7 & 70.4 & 43.7 & 67.8 & 49.2 \\
    & \textcolor{green!85}{\ding{51}} & \textcolor{green!85}{\ding{51}} & \textcolor{red!85}{\ding{55}} & \underline{84.0} & 69.6 & \underline{52.1} & \underline{35.6} & \underline{71.5} & \underline{46.4} & \underline{69.2} & \underline{50.5} \\
    & \textcolor{green!85}{\ding{51}} & \textcolor{green!85}{\ding{51}} & \textcolor{green!85}{\ding{51}} & \textbf{85.4} & \textbf{76.4} & \textbf{53.6} & \textbf{42.0} & \textbf{75.7}& \textbf{48.4} & \textbf{71.6} & \textbf{55.6} \\
    \midrule
    \multirow{4}{*}{Gemini-3.1-Pro}
    & \textcolor{red!85}{\ding{55}} & \textcolor{red!85}{\ding{55}} & \textcolor{red!85}{\ding{55}} & 92.8 & 88.0 & \underline{66.7} & 59.6 & 89.5 & 78.8 & 83.0 & 75.5 \\
    & \textcolor{green!85}{\ding{51}} & \textcolor{red!85}{\ding{55}} & \textcolor{red!85}{\ding{55}} & 92.9 & 87.2 & 66.6 & \underline{60.8} & 89.2 & 79.0 & 82.9 & 75.7 \\
    & \textcolor{green!85}{\ding{51}} & \textcolor{green!85}{\ding{51}} & \textcolor{red!85}{\ding{55}} & \underline{94.0} & \underline{90.8} & 66.2 & \underline{60.8} & \underline{90.8} & \underline{83.6} & \underline{83.7} & \underline{78.4} \\
    & \textcolor{green!85}{\ding{51}} & \textcolor{green!85}{\ding{51}} & \textcolor{green!85}{\ding{51}} & \textbf{94.1} & \textbf{91.6} & \textbf{67.7} & \textbf{63.6} & \textbf{91.2} & \textbf{84.8} & \textbf{84.3} & \textbf{80.0} \\
    \bottomrule
  \end{tabular}%
  }
  \caption{Complete results of ablation study on DocIntent components. "AG Prompt" denotes answerability-guided prompting. The Prompt-only setting enables AG Prompt while disabling AGRes and CBRb. "ANLS" denotes Average Normalized Levenshtein Similarity. Top results are \textbf{bolded} and second-best \underline{underlined}.}
  \label{tab:ablation_full}
\end{table*}

\section{Context Management Strategy}
\label{sec:context_strategy}

In agentic frameworks, context management strategies strongly affect system performance and efficiency. We explore three different context management approaches. The No Context strategy uses independent conversations for both Phase A and Phase B, retaining no historical context. The Full Context strategy retains all conversation history and intermediate restored images, providing complete reasoning history. The Text-only strategy, our approach, retains text context such as tool invocation decisions and quality assessment results but does not retain intermediate restored images. Table~\ref{tab:context_strategy} presents the experimental results of the three strategies.

The experimental results reveal that while the No Context strategy can reduce token consumption, it leads to notable performance degradation. For example, Qwen3.5-9B achieves only 40.7\% accuracy on WildChartQA under the No Context strategy, substantially lower than other strategies. We attribute this to the fact that historical context helps models make better judgments and complete tasks more effectively. Additionally, the No Context strategy may cause smaller models to misjudge during the Phase B quality comparison, leading to repeated invocations of the same tool and consequently increased token consumption. The Full Context strategy achieves high accuracy but incurs substantially higher token consumption. For instance, Qwen3.5-9B's average token consumption is 29439, 1.55 times that of the Text-only strategy. Furthermore, for less capable models, multiple images in long contexts can cause attention dispersion issues. In contrast, the Text-only strategy achieves an excellent balance between performance and efficiency, attaining the best or near-best results on most metrics while effectively controlling token consumption. Based on these observations, we chose the Text-only strategy after weighing performance and efficiency, leveraging MLLMs' image understanding capabilities for accurate decisions based on current images while retaining text context to provide necessary historical experience and avoid repeated errors.

\section{Reproducibility Details}
\label{sec:reproducibility}

The system prompts and tool strategies were developed using samples outside the evaluation sets. These samples were used only to validate the end-to-end workflow and determine answer formats, without tuning based on evaluation results.

All document images are dynamically resized while preserving their original aspect ratios, with the longest edge capped at 2,048 pixels to avoid API payload limits and out-of-memory errors. For all evaluated open- and closed-source models, we set \texttt{temperature=0.0}. Where supported, we use \texttt{top\_p=1.0}, \texttt{frequency\_penalty=0.0}, and \texttt{presence\_penalty=0.0}. The maximum output length is set to 4,096 tokens, while all other generation parameters retain their default values.

All restoration tools use their official open-source checkpoints and default inference scripts. We adapt each restoration model to DocIntent through the standardized tool interface described in Section~\ref{sec:framework}.

\section{Cost Analysis}

Table~\ref{tab:token_consumption} reports the average token consumption of different models with and without DocIntent. The additional overhead mainly arises from answerability assessment and comparison between the images before and after restoration, which support selective restoration and Comparison-Based Rollback.

For closed-source MLLMs, Table~\ref{tab:api_cost} further reports the exact API versions, official input/output token prices at the time of evaluation, and DocIntent's average token usage and per-sample cost. For example, Gemini-3.1-Pro consumes an average of 5,033 input tokens and 1,073 output tokens, corresponding to approximately \$0.0394 per sample.

\begin{table*}[t]
  \centering
  \resizebox{\textwidth}{!}{
  \begin{tabular}{l l c c c}
    \toprule
    \textbf{Model} & \textbf{API Version} & \textbf{Price (\$/1M, In/Out)} & \textbf{Avg. Tokens (In/Out)} & \textbf{Cost/Sample (\$)} \\
    \midrule
    Gemini-3.1-Pro & \texttt{gemini-3.1-pro-preview} & 4.00/18.00 & 5,033/1,073 & 0.0394 \\
    GPT-4o & \texttt{gpt-4o-2024-08-06} & 2.50/10.00 & 12,004/2,468 & 0.0547 \\
    GPT-5.2 & \texttt{gpt-5.2-2025-12-11} & 1.75/14.00 & 16,758/2,444 & 0.0635 \\
    Claude Sonnet 4.6 & \texttt{claude-sonnet-4-6} & 5.00/25.00 & 11,124/1,504 & 0.0932 \\
    Qwen3.5-Plus & \texttt{qwen3.5-plus-2026-02-15} & 0.57/3.43 & 16,814/2,220 & 0.0172 \\
    Doubao-2.0-Pro & \texttt{doubao-seed-2-0-pro-260215} & 1.37/6.86 & 6,992/933 & 0.0160 \\
    \bottomrule
  \end{tabular}
  }
  \caption{API versions, token usage, and per-sample inference costs of DocIntent for closed-source MLLMs. Prices are official input/output token prices at the time of evaluation.}
  \label{tab:api_cost}
\end{table*}

\begin{figure}[t]
  \centering
  \includegraphics[width=\columnwidth]{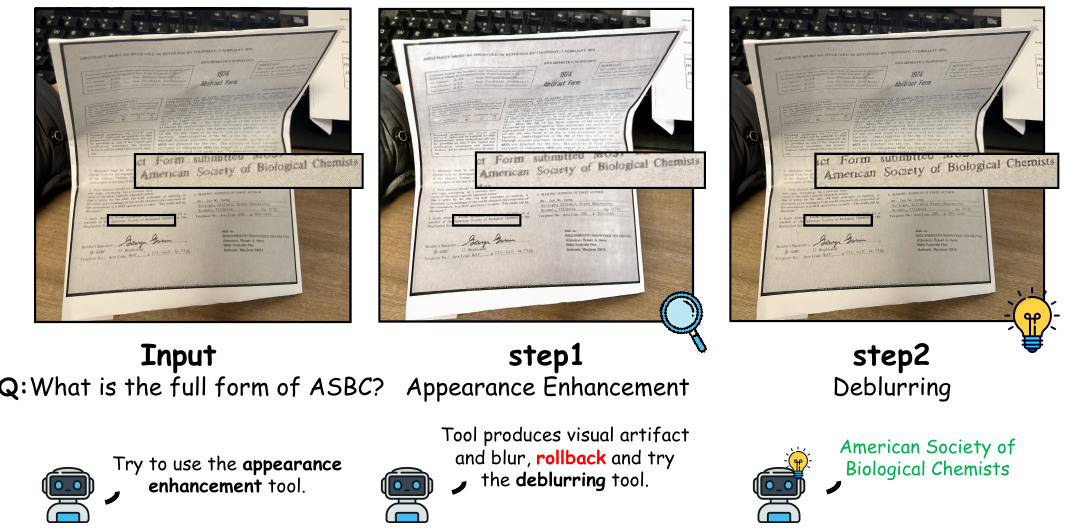}
  \caption{Rollback mechanism demonstration. Failed restoration triggers rollback and alternative attempt.}
  \label{fig:case_rollback}
\end{figure}

\begin{table*}[t]
  \centering
  \resizebox{\textwidth}{!}{%
  \begin{tabular}{l l cc cc cc cc}
    \toprule
    \multicolumn{1}{c}{\multirow{2}{*}{\textbf{Model}}} & \multicolumn{1}{c}{\multirow{2}{*}{\textbf{Tools}}} & \multicolumn{2}{c}{\textbf{WildDocVQA}} & \multicolumn{2}{c}{\textbf{WildChartQA}} & \multicolumn{2}{c}{\textbf{WildTableVQA}} & \multicolumn{2}{c}{\textbf{Average}} \\
    \cmidrule(lr){3-4} \cmidrule(lr){5-6} \cmidrule(lr){7-8} \cmidrule(lr){9-10}
    & & ANLS & Consistency & Acc. & Consistency & Acc. & Consistency & Score & Consistency \\
    \midrule
    \multirow{2}{*}{Qwen3.5-9B}
    & Group 1 & 85.4 & 76.4 & \textbf{53.6} & \textbf{42.0} & \textbf{75.7} & \textbf{48.4} & \textbf{71.6} & \textbf{55.6} \\
    & Group 2 & \textbf{86.3} & \textbf{76.8} & 52.5 & 39.6 & 73.9 & 46.8 & 70.9 & 54.4 \\
    \midrule
    \multirow{2}{*}{Gemini-3.1-Pro}
    & Group 1 & 94.1 & \textbf{91.6} & \textbf{67.7} & \textbf{63.6} & \textbf{91.2} & \textbf{84.8} & \textbf{84.3} & \textbf{80.0} \\
    & Group 2 & \textbf{94.4} & 91.2 & 67.3 & 62.8 & 90.7 & 82.8 & 84.1 & 78.9 \\
    \bottomrule
  \end{tabular}%
  }
  \caption{Performance comparison of DocIntent with different tools. "ANLS" denotes Average Normalized Levenshtein Similarity. Better results are \textbf{bolded}.}
  \label{tab:tool_backends}
\end{table*}

The cost also depends on each model's tool invocation strategy. Doubao-2.0-Pro~\cite{seedseed2} invokes tools 0.69 times per sample on average, whereas GPT-4o~\cite{hurst2024gpt} and GPT-5.2~\cite{singh2025openai} invoke tools approximately twice, indicating that model-specific decision preferences affect the final token consumption.

Overall, DocIntent trades additional inference cost for improvements in document VQA performance and consistency. In practical deployment, the backbone model can be selected according to the desired balance between performance and cost.

\begin{table}[t]
  \centering
  \small
  \begin{tabular}{l l c}
    \toprule
    \textbf{Model} & \textbf{Method} & \textbf{ANLS} \\
    \midrule
    \multirow{2}{*}{Qwen3.5-9B}
    & Direct Answer & \textbf{95.3} \\
    & DocIntent & 94.7 \\
    \midrule
    \multirow{2}{*}{Gemini-3.1-Pro}
    & Direct Answer & \textbf{95.6} \\
    & DocIntent & 95.4 \\
    \bottomrule
  \end{tabular}
  \caption{Performance on the clean DocVQA~\cite{mathew2021docvqa} test set. Better results within each model are \textbf{bolded}.}
  \label{tab:clean_docvqa}
\end{table}

\begin{table*}[t]
  \centering
  \resizebox{\textwidth}{!}{%
  \begin{tabular}{l l cc cc cc cc}
    \toprule
    \multicolumn{1}{c}{\multirow{2}{*}{\textbf{Model}}} & \multicolumn{1}{c}{\multirow{2}{*}{\textbf{Method}}} & \multicolumn{2}{c}{\textbf{WildDocVQA}} & \multicolumn{2}{c}{\textbf{WildChartQA}} & \multicolumn{2}{c}{\textbf{WildTableVQA}} & \multicolumn{2}{c}{\textbf{Average}} \\
    \cmidrule(lr){3-4} \cmidrule(lr){5-6} \cmidrule(lr){7-8} \cmidrule(lr){9-10}
    & & ANLS & Cons. & Acc. & Cons. & Acc. & Cons. & Score & Cons. \\
    \midrule
    \multirow{2}{*}{Qwen3.5-9B}
    & Direct Answer & $82.5 \pm \text{0.1}$ & $70.8 \pm \text{0.1}$ & $51.9 \pm \text{0.3}$ & $33.2 \pm \text{0.2}$ & $71.1 \pm \text{0.1}$ & $45.2 \pm \text{0.2}$ & $68.5 \pm 0.1$ & $49.7 \pm 0.1$ \\
    & DocIntent & $\mathbf{85.4 \pm \text{0.2}}$ & $\mathbf{76.4 \pm \text{0.1}}$ & $\mathbf{53.6 \pm \text{0.4}}$ & $\mathbf{42.0 \pm \text{0.4}}$ & $\mathbf{75.7 \pm \text{0.2}}$ & $\mathbf{48.4 \pm \text{0.1}}$ & $\mathbf{71.6 \pm 0.3}$ & $\mathbf{55.6 \pm 0.2}$ \\
    \midrule
    \multirow{2}{*}{Gemini-3.1-Pro}
    & Direct Answer & $92.8 \pm \text{0.3}$ & $88.0 \pm \text{0.2}$ & $66.7 \pm \text{0.5}$ & $59.6 \pm \text{0.6}$ & $89.5 \pm \text{0.2}$ & $78.8 \pm \text{0.4}$ & $83.0 \pm 0.3$ & $75.5 \pm 0.5$ \\
    & DocIntent & $\mathbf{94.1 \pm \text{0.3}}$ & $\mathbf{91.6 \pm \text{0.2}}$ & $\mathbf{67.7 \pm \text{0.6}}$ & $\mathbf{63.6 \pm \text{0.6}}$ & $\mathbf{91.2 \pm \text{0.2}}$ & $\mathbf{84.8 \pm \text{0.2}}$ & $\mathbf{84.3 \pm 0.4}$ & $\mathbf{80.0 \pm 0.4}$ \\
    \bottomrule
  \end{tabular}%
  }
  \caption{Mean and standard deviation over three runs on the WildDoc~\cite{wang2025wilddoc} benchmark. Better results within each model are \textbf{bolded}.}
  \label{tab:repeated_runs}
\end{table*}

\begin{table}[t]
  \centering
  \resizebox{\columnwidth}{!}{%
  \begin{tabular}{l c ccc c}
    \toprule
    \textbf{Model} & \textbf{Avg. Tools} & \textbf{WildDocVQA} & \textbf{WildChartQA} & \textbf{WildTableVQA} & \textbf{Average} \\
    \midrule
    Qwen3.5-9B & 0.00 & 4943 & 4858 & 6137 & 5313 \\
    w/ DocIntent & 1.69 & 17967 & 19611 & 19518 & 19032 \\
    \midrule
    Gemini-3.1-Pro & 0.00 & 2019 & 2226 & 2882 & 2376 \\
    w/ DocIntent & 1.16 & 5817 & 6791 & 5711 & 6106 \\
    \midrule
    Qwen3.5-27B & 0.00 & 4839 & 4658 & 5756 & 5084 \\
    w/ DocIntent & 1.42 & 17836 & 16523 & 16528 & 16962 \\
    \midrule
    GPT-4o & 0.00 & 1710 & 2023 & 2553 & 2095 \\
    w/ DocIntent & 2.02 & 13894 & 15448 & 14076 & 14473 \\
    \midrule
    Qwen3.5-Plus & 0.00 & 3440 & 4464 & 4530 & 4145 \\
    w/ DocIntent & 1.87 & 17347 & 18641 & 21113 & 19034 \\
    \midrule
    Claude Sonnet 4.6 & 0.00 & 3227 & 3211 & 3670 & 3369 \\
    w/ DocIntent & 1.84 & 12264 & 13276 & 12345 & 12628 \\
    \midrule
    Doubao-2.0-Pro & 0.00 & 3342 & 3964 & 4272 & 3859 \\
    w/ DocIntent & 0.69 & 6537 & 8086 & 9152 & 7925 \\
    \midrule
    GPT-5.2 & 0.00 & 1867 & 2138 & 3217 & 2407 \\
    w/ DocIntent & 2.00 & 16204 & 18906 & 22495 & 19202 \\
    \bottomrule
  \end{tabular}%
  }
  \caption{Average token consumption per sample w/ and w/o DocIntent across different models and tasks.}
  \label{tab:token_consumption}
\end{table}

\section{Complete Experimental Results}
\label{sec:complete_results}

This section provides the complete experimental results for both the strategy comparison (Table~\ref{tab:strategy_comparison_full}) and ablation study (Table~\ref{tab:ablation_full}). These tables present detailed performance metrics across all three subtasks of the WildDoc benchmark: WildDocVQA, WildChartQA, and WildTableVQA.

\section{Impact of Tool Selection}
\label{sec:tool_selection}

To analyze the impact of different tool choices on DocIntent, we conduct comparative experiments with two tool backend configurations. Group 1 employs DocRes~\cite{zhang2024docres} for deblurring, dewarping, deshadowing, and appearance enhancement, combined with UniDemoiré~\cite{yang2025unidemoire} for demoiréing. Group 2 uses NAF-DPM~\cite{cicchetti2024naf} for deblurring, BGShadowNet~\cite{zhang2023document} for deshadowing, DocRes~\cite{zhang2024docres} for dewarping and appearance enhancement, and UniDemoiré~\cite{yang2025unidemoire} for demoiréing.

Table~\ref{tab:tool_backends} presents the experimental results. For Qwen3.5-9B, Group 1 achieves a 71.6\% average score and 55.6\% consistency, while Group 2 obtains 70.9\% and 54.4\%. For Gemini-3.1-Pro, Group 1 achieves an 84.3\% average score and 80.0\% consistency, while Group 2 achieves 84.1\% and 78.9\%. The two configurations show similar performance levels across both models. These results indicate that DocIntent achieves effective performance improvements regardless of tool backend choices. The answerability-guided restoration strategy and comparison-based rollback mechanism can effectively coordinate different restoration tool combinations. Based on overall performance, we adopt the Group 1 configuration in our main experiments. Notably, our contribution lies in effectively orchestrating restoration tools to enhance performance rather than improving existing restoration tools.

\section{Human Expert Annotation Details}
\label{sec:expert_annotation}

We recruited 40 experts with document image processing experience to construct the Human Expert baseline. All participants are graduate students with relevant experience in document restoration. Before the annotation process, each expert received detailed task guidelines, including an introduction to degradation types in the WildDoc dataset, functional descriptions of all restoration tools, and several pre-annotated examples to ensure they fully understood the task requirements.

All samples were evenly distributed among the 40 experts. Each sample was assigned to exactly one expert, who developed its restoration strategy.

The annotation task required experts to manually identify degradation types for each test sample and select appropriate tools to optimize overall document readability. Importantly, experts were not provided with the associated questions during annotation, focusing solely on achieving the best overall document readability. The annotation task was completed over a span of 20 days, requiring approximately 10 hours of actual workload per expert, with a total cost of \$1,750. All experts participated with informed consent and understood that their annotation data would be used for academic research purposes.

For the question-aware Human Expert baseline, experts were additionally provided with the corresponding questions and instructed to locate question-relevant evidence and select tools to improve its readability. Experts could skip restoration when the relevant evidence was already clearly visible. Any orientation correction selected by an expert was applied immediately, allowing subsequent restoration operations to be planned based on the corrected document.

\section{Repeated-Run Evaluation}

To evaluate the stability of our results, we repeated the WildDoc evaluation three times for Qwen3.5-9B and Gemini-3.1-Pro with and without DocIntent. Table~\ref{tab:repeated_runs} reports the mean and standard deviation across the three runs. DocIntent consistently improves both the average score and consistency for both models, while the small standard deviations demonstrate stable performance across repeated runs.

\section{Performance on Clean Documents}

To evaluate whether DocIntent affects question-answering performance on clean documents, we test Qwen3.5-9B and Gemini-3.1-Pro on the DocVQA~\cite{mathew2021docvqa} test set. The results in Table~\ref{tab:clean_docvqa} show that DocIntent largely preserves the models' performance on clean data without causing a notable degradation.

\section{Complete Prompt Templates}
\label{sec:prompts}

We provide complete prompt templates for all three WildDoc subtasks: WildDocVQA, WildChartQA, and WildTableVQA. These include: (1) DocIntent prompts containing task instructions, answerability-guided restoration strategy, tool descriptions, and comparison-based rollback mechanism; (2) baseline prompts for direct answering in comparison experiments; (3) Phase B comparison prompt for quality assessment and rollback decisions; and (4) tool descriptions provided to the MLLM.

To ensure fair comparison, DocIntent and baseline prompts for the same task share identical answer format rules, including extraction principles, capitalization preservation, number formatting, and unit symbol handling. The only difference lies in that DocIntent prompts include answerability-guided restoration strategy and tool descriptions, while baseline prompts contain only task instructions and answer format rules without any restoration tools or strategy guidance. This design ensures that performance differences stem from the methods themselves rather than answer format variations.

DocIntent prompts contain four key components: (1) task instructions defining answer format requirements, (2) answerability-guided restoration strategy that explicitly instructs the MLLM to select the most severe degradation type whose tool remains available when multiple degradations coexist, (3) descriptions of all available tools covering image quality restoration and orientation correction, and (4) mandatory comparison and review mechanism after each tool execution, enforced through automatic message insertion and dynamic tool list adjustment. These prompts guide the MLLM to complete tasks through answerability-guided restoration with comparison-based rollback.

\begin{figure*}[t]
  \centering
  \includegraphics[width=\textwidth]{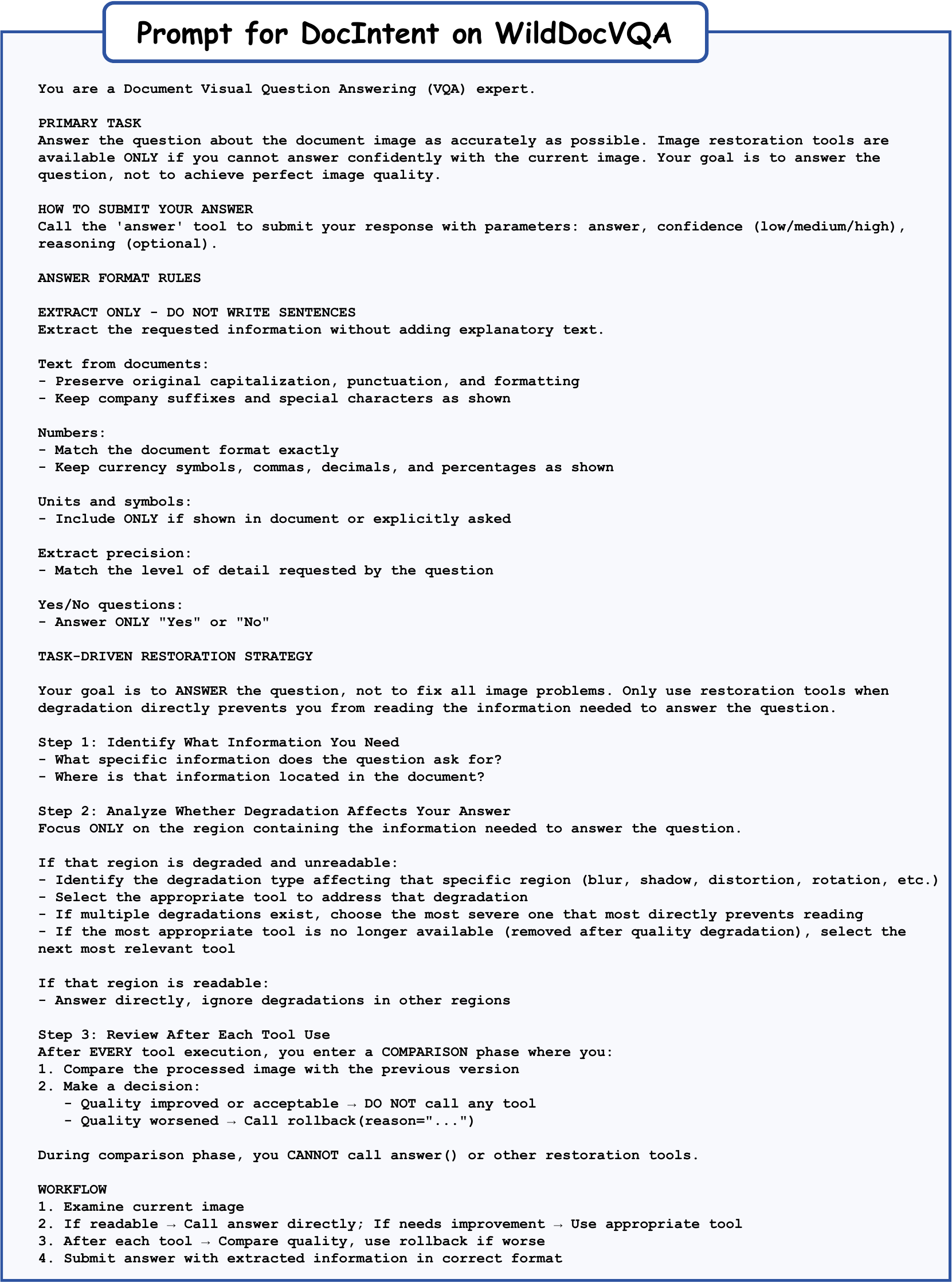}
  \caption{Prompt for DocIntent on WildDocVQA task.}
  \label{prompt:docvqa}
\end{figure*}

\begin{figure*}[t]
  \centering
  \includegraphics[width=\textwidth]{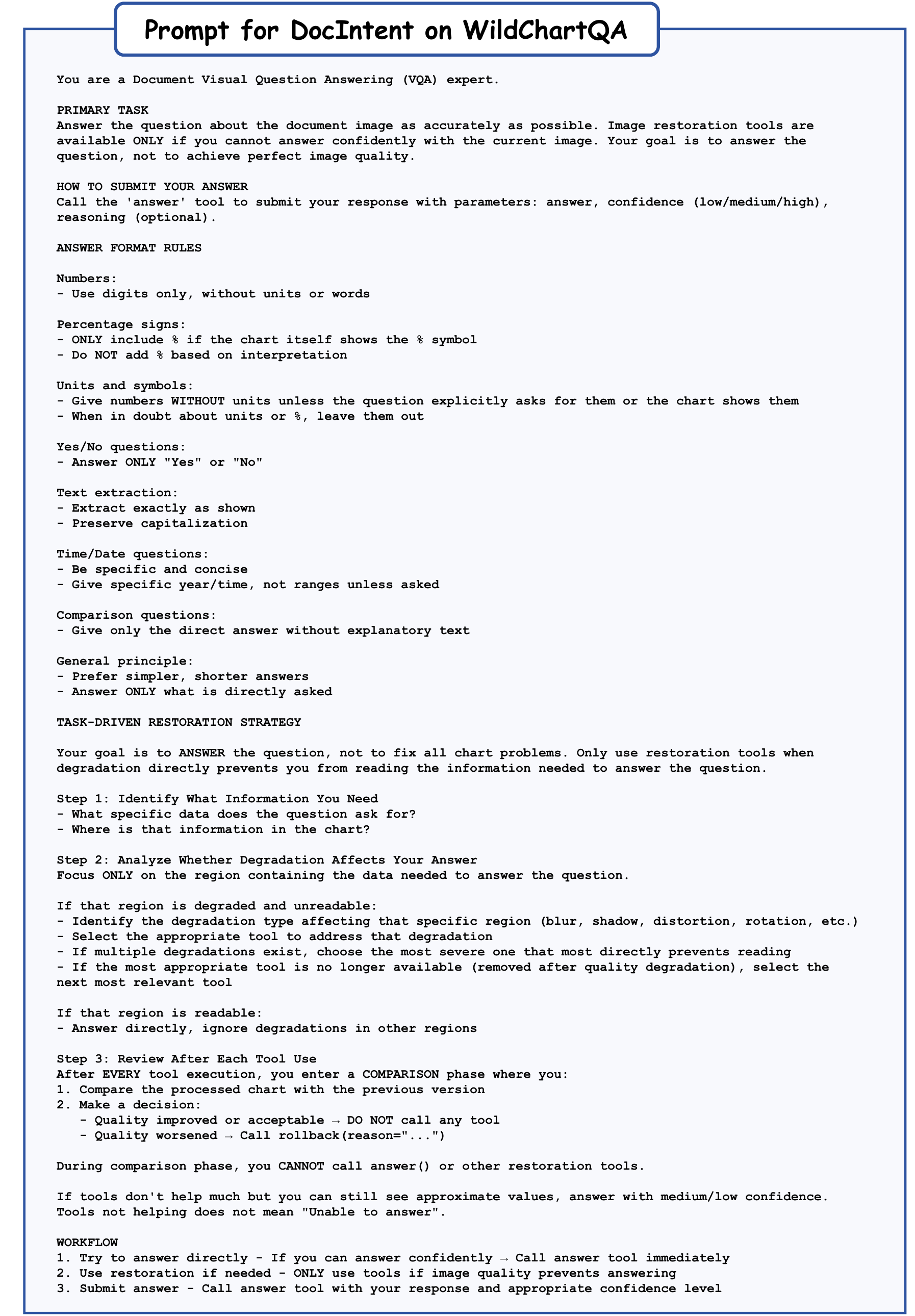}
  \caption{Prompt for DocIntent on WildChartQA task.}
  \label{prompt:chartqa}
\end{figure*}

\begin{figure*}[t]
  \centering
  \includegraphics[width=\textwidth]{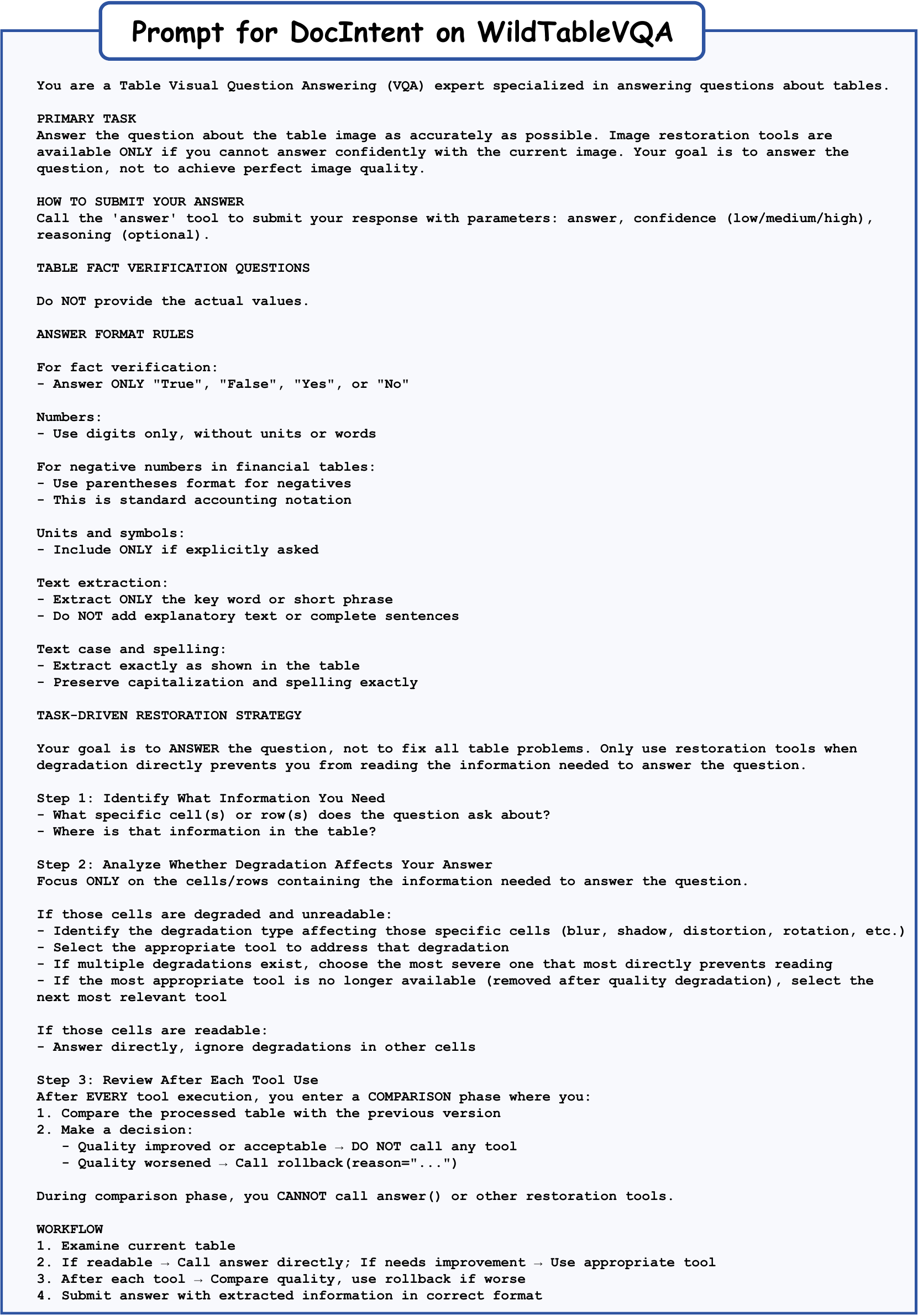}
  \caption{Prompt for DocIntent on WildTableVQA task.}
  \label{prompt:tablevqa}
\end{figure*}

\begin{figure*}[t]
  \centering
  \includegraphics[width=\textwidth]{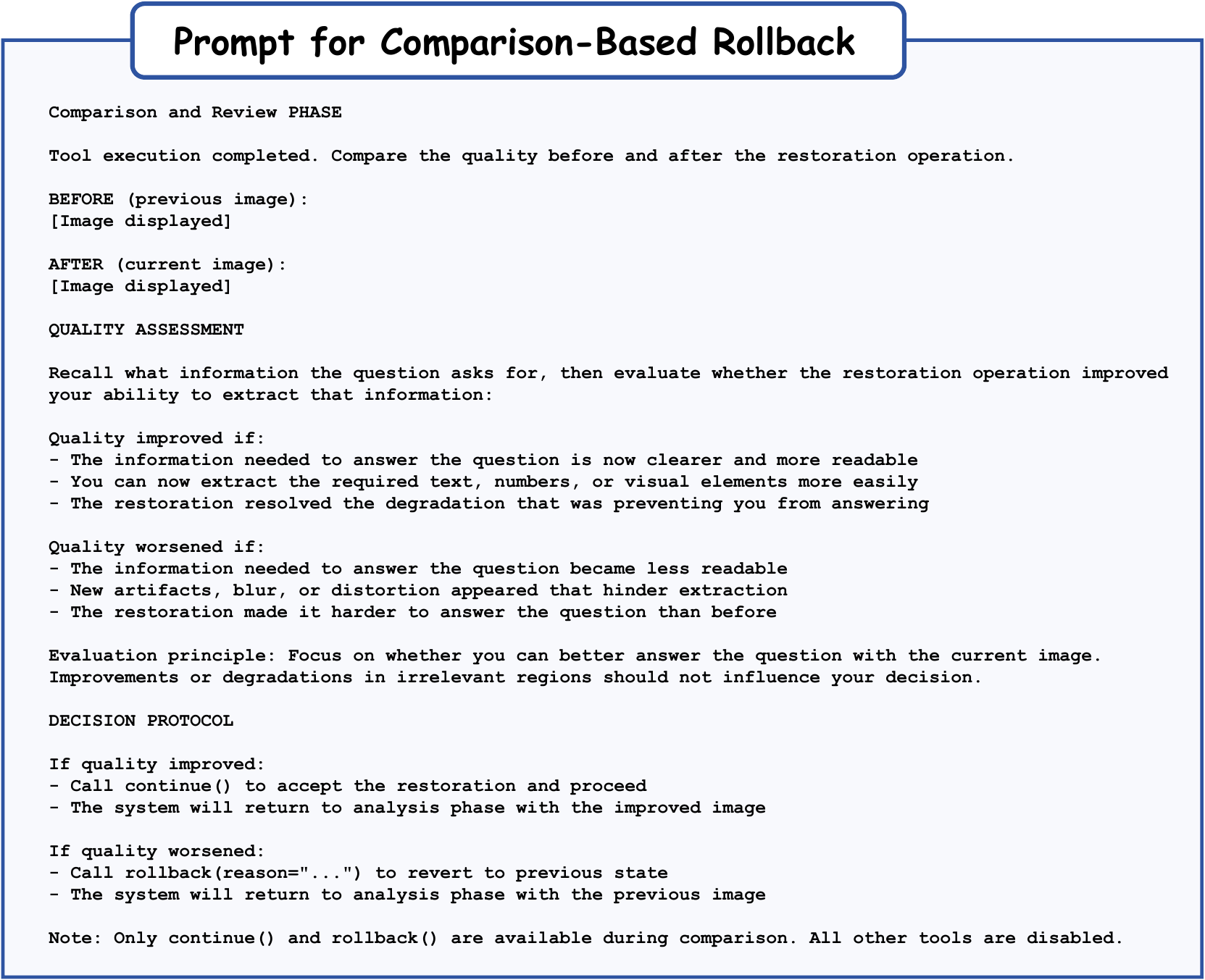}
  \caption{Prompt for Comparison-Based Rollback.}
  \label{prompt:rollback}
\end{figure*}

\begin{figure*}[t]
  \centering
  \includegraphics[width=\textwidth]{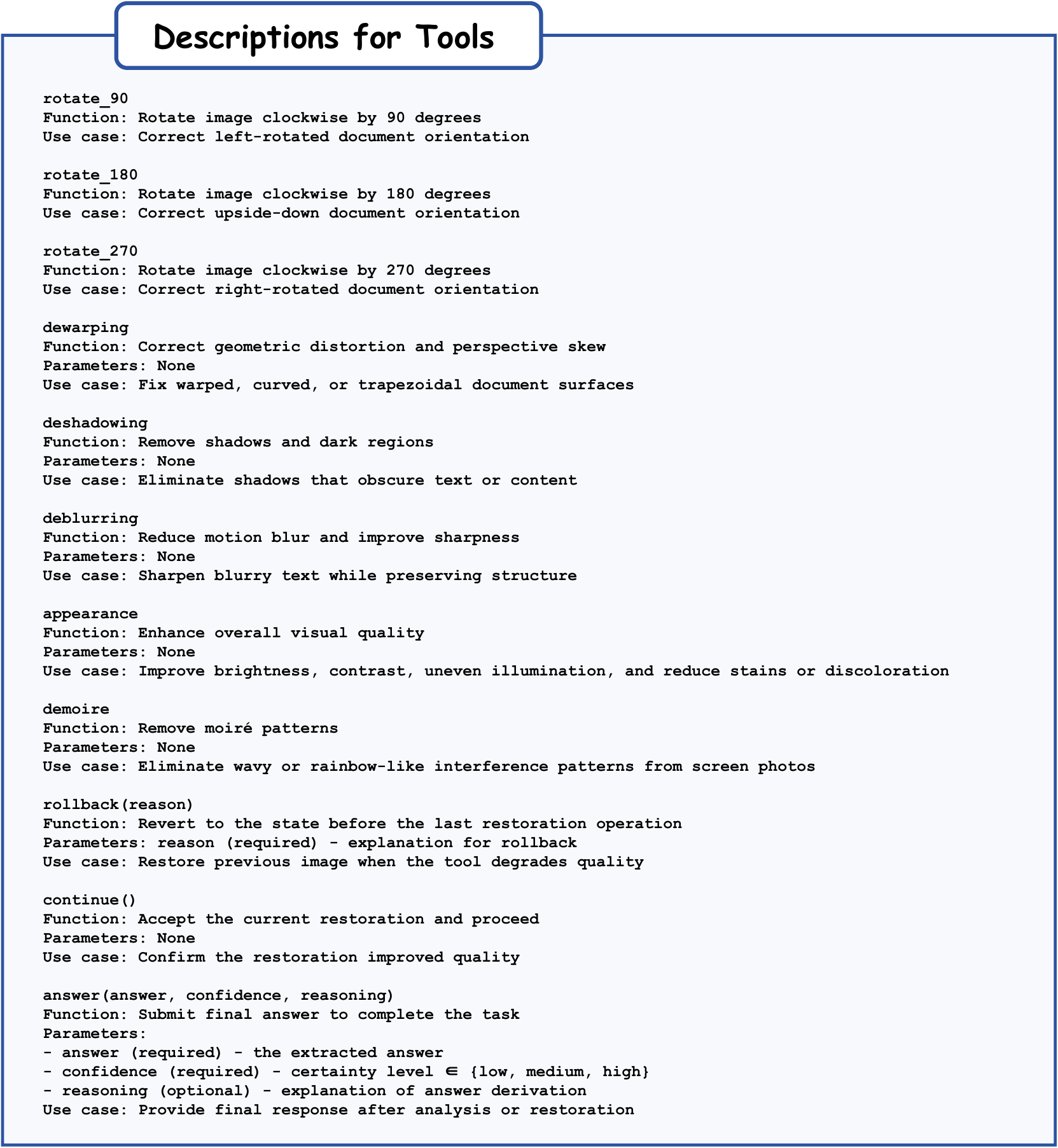}
  \caption{Tools descriptions.}
  \label{prompt:tools}
\end{figure*}

\begin{figure*}[t]
  \centering
  \includegraphics[width=\textwidth]{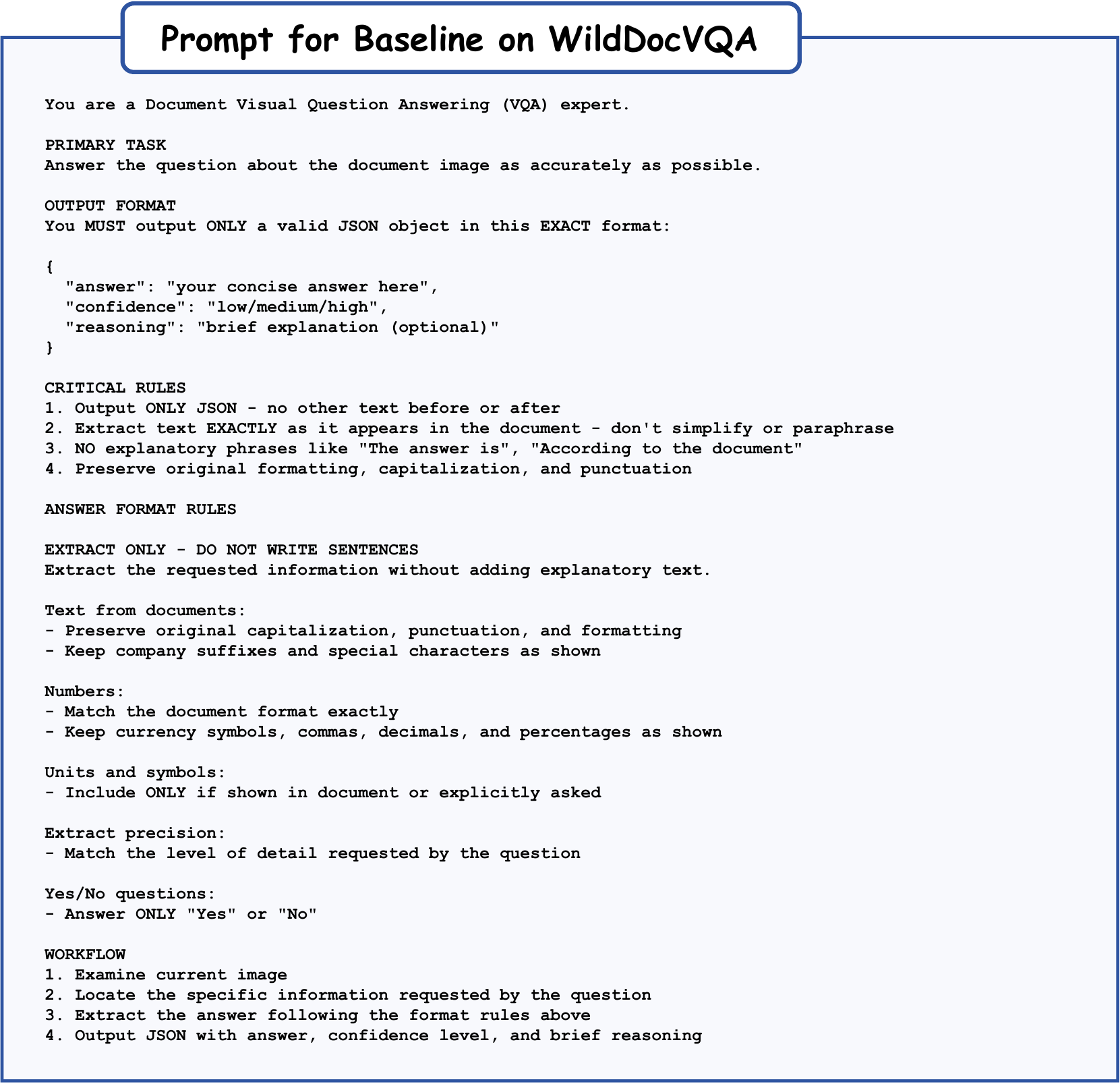}
  \caption{Prompt for Baseline on WildDocVQA task.}
  \label{prompt:docvqa_baseline}
\end{figure*}

\begin{figure*}[t]
  \centering
  \includegraphics[width=\textwidth]{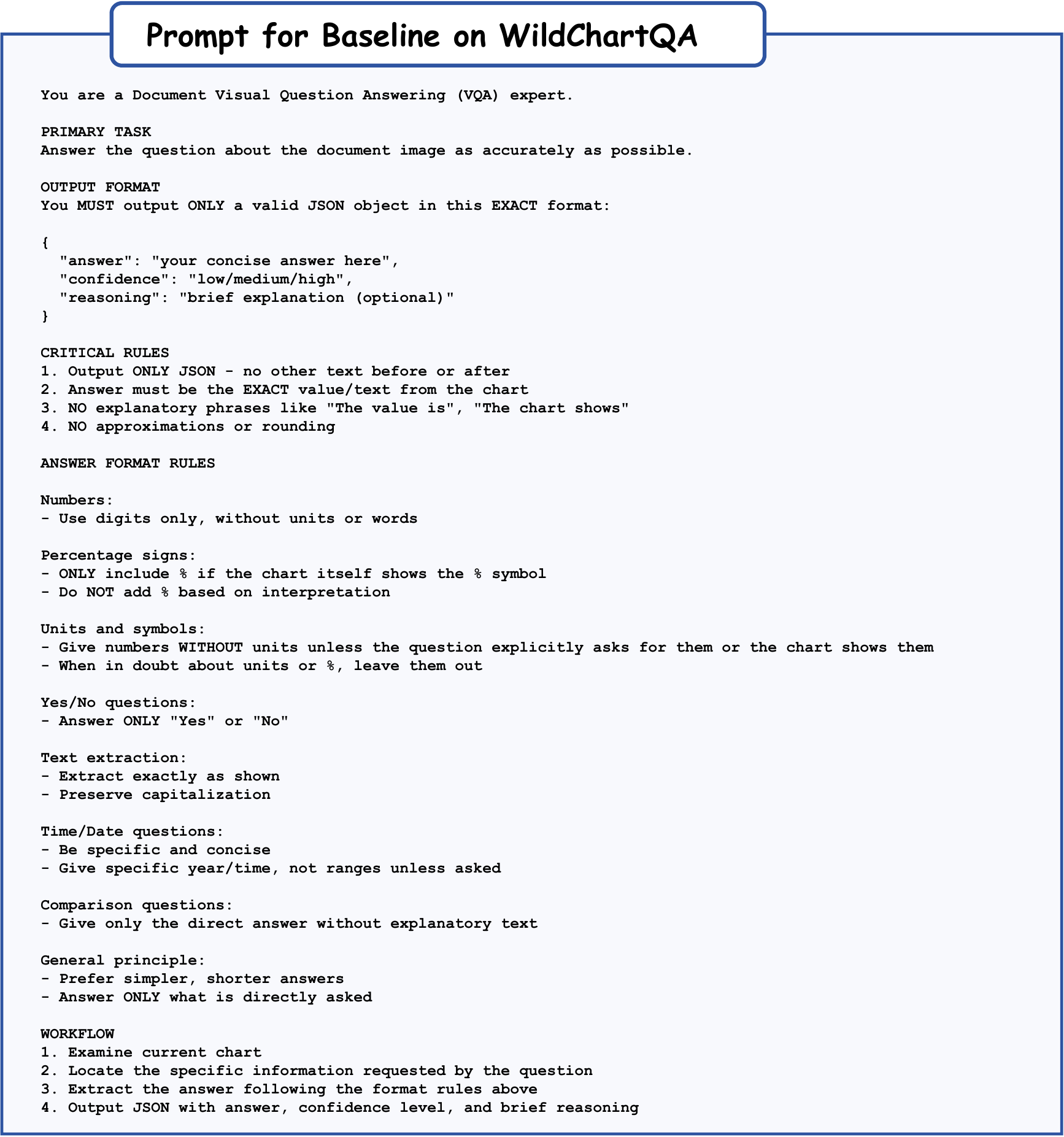}
  \caption{Prompt for Baseline on WildChartQA task.}
  \label{prompt:chartqa_baseline}
\end{figure*}

\begin{figure*}[t]
  \centering
  \includegraphics[width=\textwidth]{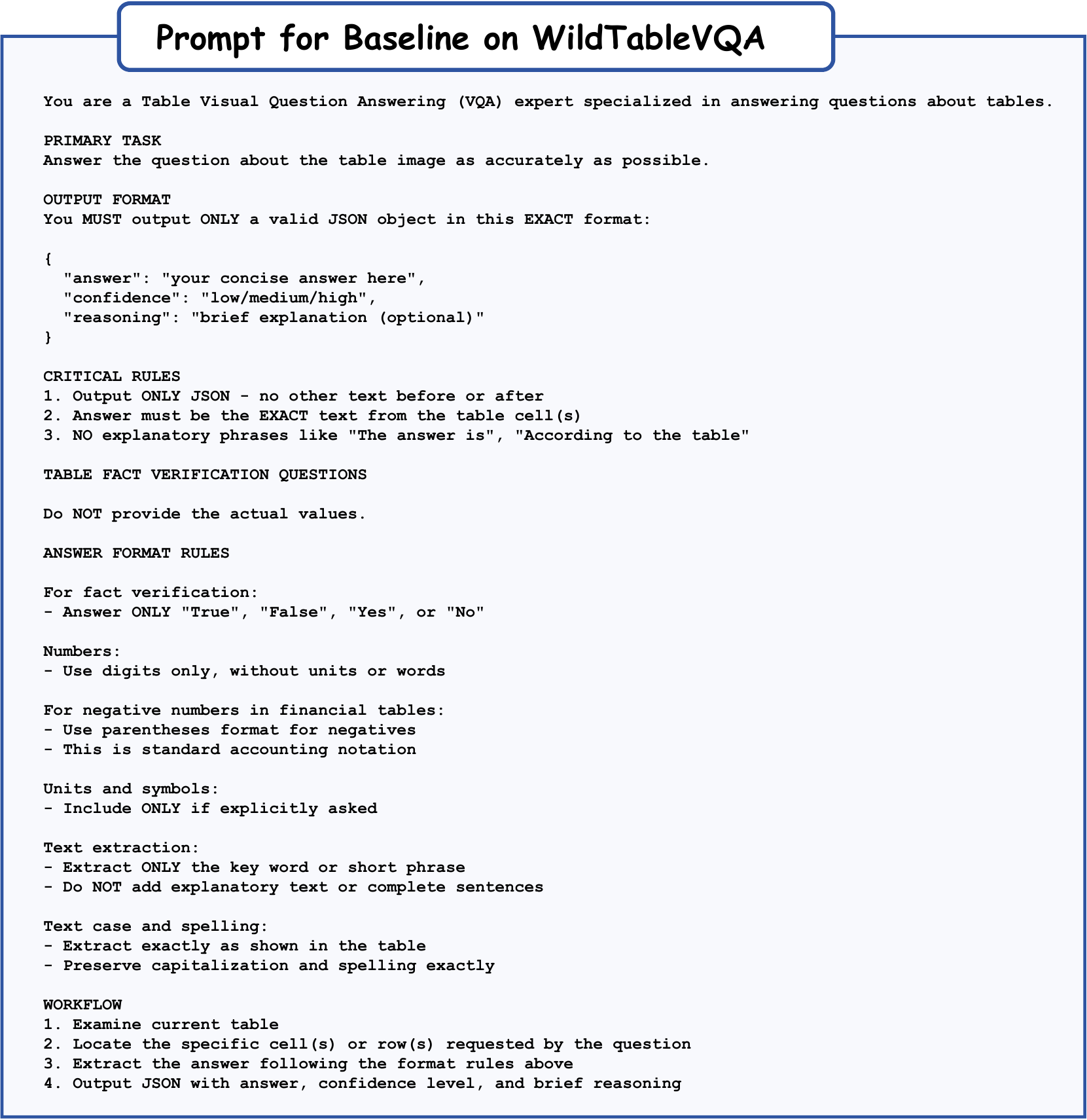}
  \caption{Prompt for Baseline on WildTableVQA task.}
  \label{prompt:tablevqa_baseline}
\end{figure*}

\end{document}